\documentclass[10pt,twocolumn,letterpaper]{article}
\usepackage{url}            
\usepackage{xcolor}         

\usepackage{times}
\usepackage{microtype} 
\usepackage{graphicx} 
\usepackage{wrapfig}
\usepackage{balance} 

\usepackage{algorithm}
\usepackage{algorithmic}

\usepackage{helvet}
\usepackage{courier}

\usepackage{makecell}
\usepackage{graphicx}
\usepackage{color}
\usepackage{amsfonts}
\usepackage{amsmath}
\usepackage{amssymb}

\usepackage{bm}

\usepackage{multirow}
\usepackage{booktabs}

\def\ie{\mbox{\textit{i.e.}, }}
\def\eg{\mbox{\textit{e.g.}, }}

\def\mA{{\mathcal A}}

\def\mD{{\mathcal D}}

\def\mF{{\mathcal F}}

\def\mL{{\mathcal L}}

\def\mP{{\mathcal P}}

\def\mR{{\mathcal R}}

\def\mT{{\mathcal T}}

\DeclareMathAlphabet\mathbfcal{OMS}{cmsy}{b}{n}

\def\0{{\bf 0}}
\def\1{{\bf 1}}

\def\bI{{\bm{I}}}

\def\bZ{{\bm{Z}}}

\def\bff{{\bm f}}
\def\bg{{\bm g}}

\def\bmm{{\bm m}}

\def\bo{{\bm o}}

\def\bx{{\bm x}}

\usepackage{ntheorem}

\newtheorem*{*thm}{Theorem}

\newtheorem*{*lemma}{Lemma}

\usepackage{stackengine}
\usepackage{tabularx}
\usepackage{threeparttable}
\usepackage{arydshln}

\def\red{\textcolor{black}}

\definecolor{mygray}{gray}{.9}

\usepackage{pifont}

\usepackage{adjustbox}
\newcommand{\name}{0}
\newcommand{\h}{0}
\newcommand{\w}{0.15}

\newlength \g

\usepackage{enumitem}

\newcommand{\thickhline}{%
	\noalign {\ifnum 0=`}\fi \hrule height 1pt
	\futurelet \reserved@a \@xhline
}

\usepackage{iccv}
\usepackage{times}
\usepackage{epsfig}
\usepackage{graphicx}
\usepackage{amsmath}
\usepackage{amssymb}

\usepackage{multirow, booktabs}

\usepackage[pagebackref=true,breaklinks=true,letterpaper=true,colorlinks,bookmarks=false]{hyperref}

\iccvfinalcopy 

\def\iccvPaperID{6944} 

\ificcvfinal\fi
\usepackage{subfiles}

\begin{document}

\title{Learning Task-Oriented Flows to Mutually Guide Feature Alignment in Synthesized and Real Video Denoising}

\author{%
  Jiezhang Cao$^{1}$, Qin Wang$^{1}$, Jingyun Liang$^{1}$, Yulun Zhang$^{1}$, Kai Zhang$^{1}$, Radu Timofte$^{1,2}$, Luc Van Gool$^{1,3}$\\
  $^{1}$ETH Z\"urich \quad~~$^{2}$University of Wurzburg  \quad~~$^{3}$KU Leuven \\
    \url{https://github.com/caojiezhang/ReViD}\\
}

\maketitle
\ificcvfinal\thispagestyle{empty}\fi

\begin{abstract}

Video denoising aims at removing noise from videos to recover clean ones. Some existing works show that optical flow can help the denoising by exploiting the additional spatial-temporal clues from nearby frames. However, the flow estimation itself is also sensitive to noise, and can be unusable under large noise levels. To this end, we propose a new multi-scale refined optical flow-guided video denoising method, which is more robust to different noise levels. 
Our method mainly consists of a denoising-oriented flow refinement (DFR) module and a flow-guided mutual denoising propagation (FMDP) module. Unlike previous works that directly use off-the-shelf flow solutions, DFR first learns robust multi-scale optical flows, and FMDP makes use of the flow guidance by progressively introducing and refining more flow information from low resolution to high resolution. Together with real noise degradation synthesis, the proposed multi-scale flow guided denoising network achieves state-of-the-art performance on both synthetic Gaussian denoising and real video denoising. The codes will be made publicly available.

\end{abstract}

\vspace{-3mm}
\section{Introduction}
\label{sec:intro}
\vspace{-2mm}
Video denoising, with the aim of reducing the noise from a video to recover a clean video, has drawn increasing attention in the low-level computer vision community \cite{tassano2019dvdnet,tassano2020fastdvdnet,vaksman2021pacnet,davy2018vnlnet,chan2022basicvsrpp2,lee2021restore,maggioni2021efficient,huang2022neural}.
Compared with image denoising, video denoising remains a large underexplored domain.
With the advance of deep learning \cite{ren2021adaptive,zheng2021deep,zamir2021multi}, deep neural networks (DNNs) \cite{vaksman2021pacnet,tassano2020fastdvdnet,sheth2021unsupervised} have become the dominant approach for video denoising. 
Recently, the importance of optical flow has been exploited in some video restoration methods \cite{wang2019edvr,liang2022vrt,li2022flornn,chan2021basicvsrpp} as it captures temporal motion information across frames in feature alignment and propagation.

\begin{figure}[t]
    \begin{center}
    \includegraphics[width=1\linewidth]{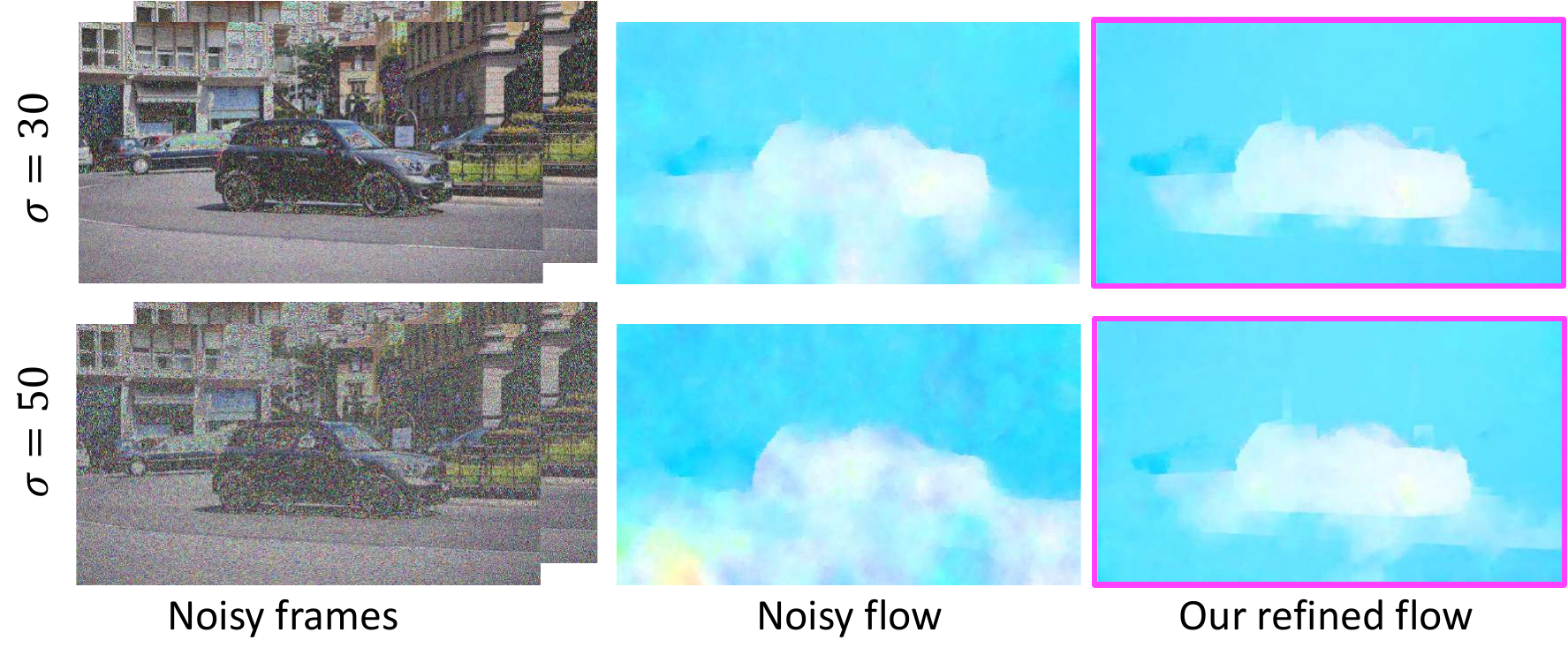}
    \vspace{-2mm}
    \includegraphics[width=1.03\linewidth]{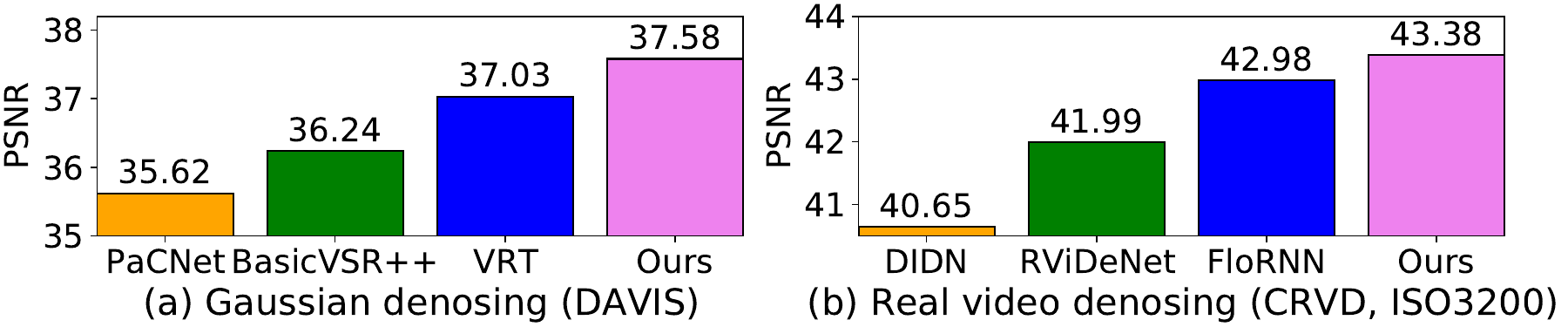}
    \end{center}
    \vspace{-1mm}
    \caption{Effect of Noise on flow (above) and comparisons with previous SOTA on both synthesized and real denoising (blow). Our method outperforms all methods with a large margin.
    }
    \label{fig:noise_flow}
    \vspace{-6mm}
\end{figure}

However, the quality of the optical flow is highly influenced by noise~\cite{xue2019toflow}, especially for high level of noise, as shown in Figure \ref{fig:noise_flow}.
Many existing flow-based video denoising methods \cite{chan2022basicvsrpp2,liang2022vrt,li2022flornn} neglect to explicitly optimize the flow estimation networks, and directly use off-the-shelf flows in the feature alignment and propagation.
As a result, the inaccurate flow guidance may lead to misleading feature alignment and feature warping error, and thus leads to poor denoising performance.
Such inaccuracy and influence can propagate across different layers.
Furthermore, the second-order propagation used in the popular BasicVSR++ \cite{chan2022basicvsrpp2} may lead to further error accumulation because of multi-step propagation. 
Therefore, how to learn robust flow estimation in the presence of noise is an unanswered key question, especially when the noise type is unknown. 


\begin{figure*}[t]
    \begin{center}
    \vspace{-4mm}
    \includegraphics[width=0.99\linewidth]{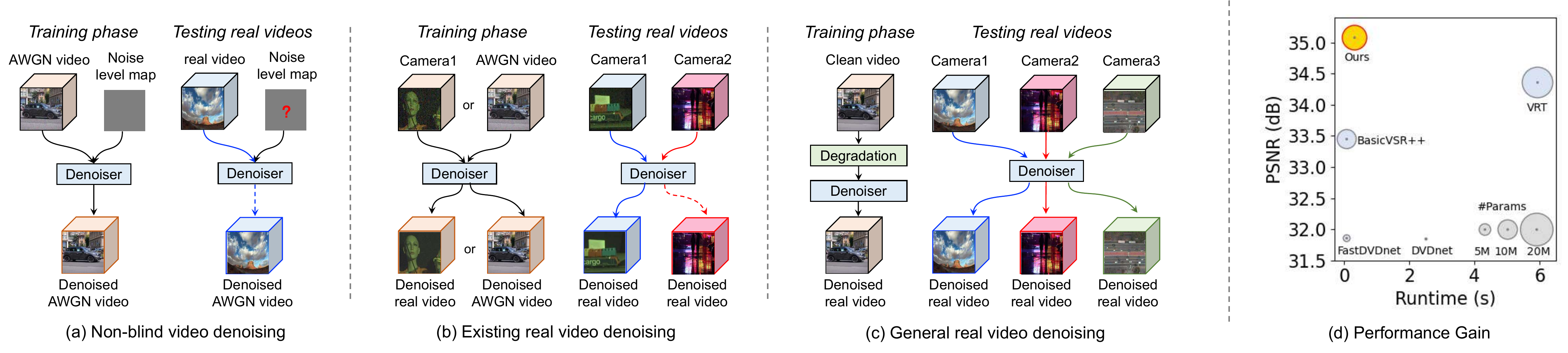}
    \end{center}
    \vspace{-4mm}
    \caption{Discussion on the difference of existing video denoising setups. (a) Non-blind denoising methods take an AWGN video and its noise as input to synthesize a clean video. (b) Existing real denoising methods aim to map a noisy video to a clean video without inputting the noise level. When training a model with noisy videos from a specific camera, it has poor performance~(marked by the dotted line) on another camera. (c) 
    We first synthesizes different kinds of noisy videos with the degradation models, and then generalize well on different real-world videos. 
    (d) Our method outperforms existing state-of-the-art video denoising methods and maintains good efficiency.
    }
    \label{fig:diff}
    \vspace{-0.5cm}
\end{figure*}

To address these, we design a new architecture to better exploit the flow guidance for video denoising.
The new architecture progressively introduces and refines more flow information from low resolution to high resolution. 
Our network consists of multiple scales, each of which has a denoising-oriented flow refinement (DFR) module and a flow-guided mutual denoising propagation (FMDP) module. 
Without directly using off-the-shelf flows, DFR optimizes the flow estimation networks by minimizing the loss between the noisy flows and clean flows.
In this way, DFR is able to reduce the sensitivity of the flow estimation networks to different levels of noise.
The FMDP module uses the refined flows to mutually guide the forward and backward propagation in each scale. Based on the refined flows, this module learn diverse offsets to discover more meaningful and relevant features.


The contributions can be summarized as follows:
\vspace{-0.2cm}
\begin{itemize} 
    \item We design a simple-but-effective video denoising architecture. Our proposed method achieves the state-of-the-art performance on both additive white Gaussian denoising tasks and real-world video denoising tasks. Moreover, our model achieves faster runtime than Transformer-based methods.


    \vspace{-0.2cm}
    \item We propose a denoising-oriented flow refinement (DFR) module and enables the flow estimation networks to be robust to different noise levels.
    Based on the refined flows, we also propose a flow-guided mutual denoising propagation (FMDP) module in which more meaningful and relevant texture can be exploited and they can be mutually propagated and aligned in different scales.

    \vspace{-0.2cm}
    \item We propose a new real video degradation model by simulating the real-world noise. The synthesized noisy videos can cover well a large range of real-world distribution. In addition, we collect a new real-world noisy videos dataset comprising diverse real-world noises. Our dataset can be utilized as a benchmark for evaluating the performance of real video denoising methods.
    
\end{itemize}

\begin{figure*}
  \begin{center}
  \vspace{-2mm}
  \includegraphics[width=1\linewidth]{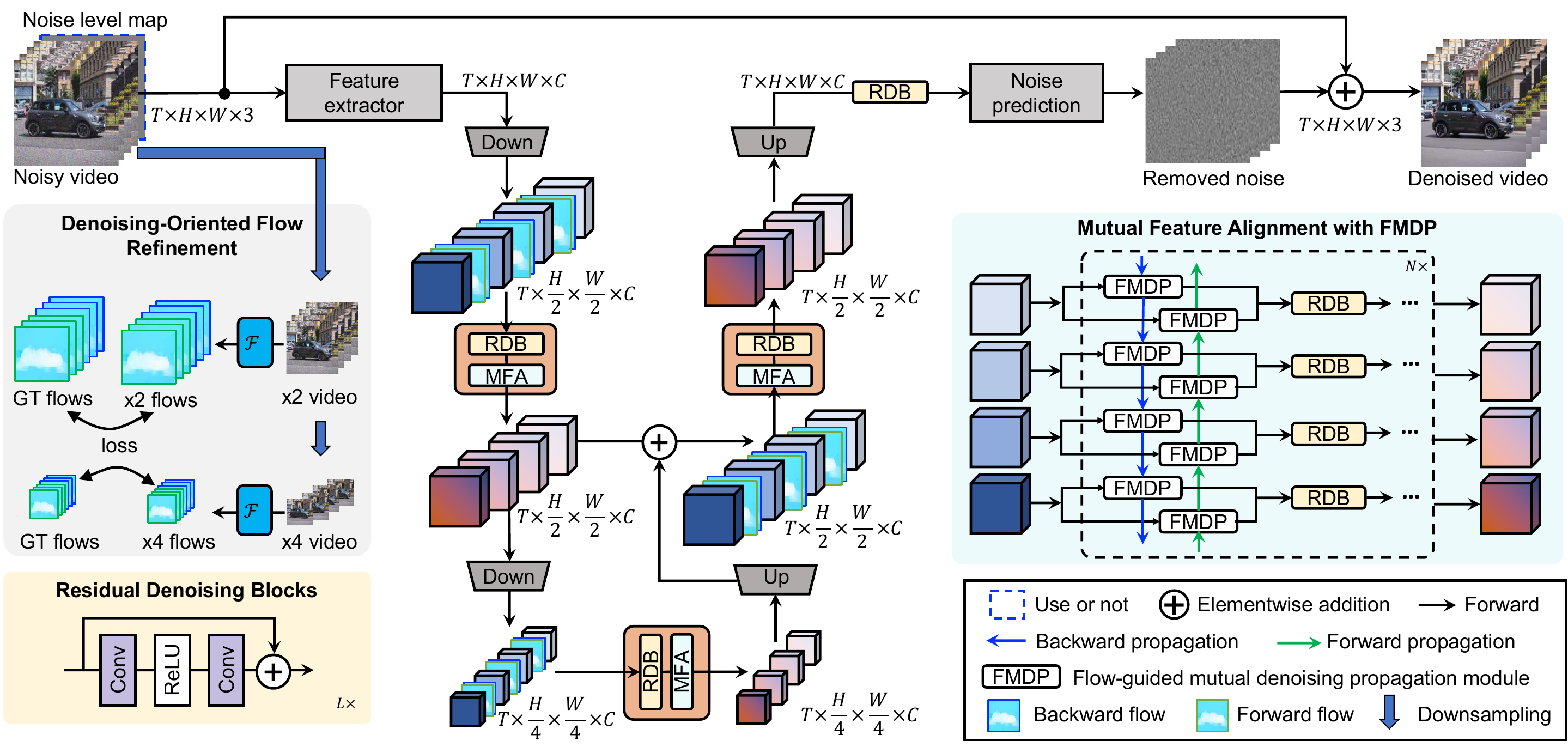}
  \end{center}
  \caption{The architecture of the proposed multi-scale recurrent network. Our network is motivated by video noise properties. For non-blind video denoising, we take the noisy video and noise level map as an input. For real video denoising, we feed the noisy video augmented by our degradation models to train the network. At each scale, we use the denoising-oriented flow refinement (DFR) module to optimize flows and then use the flow-guided mutual denoising propagation (FMDP) to mutually guide the forward and backward propagation. }
  \label{fig:arch}
\end{figure*}

\vspace{-3mm}
\section{Related Work}
\label{sec:related_work}

\vspace{-1mm}
\noindent\textbf{Image denoising}
aims to reduce noise from a noisy image \cite{kim2021noise2score,fu2021unfolding,luo2021functional,bodrito2021trainable}.
Early methods~\cite{dabov2007bm3d, lebrun2013nlb} 
depend on specific priors and hand-tuned parameters in the optimization.
To address this, recent methods exploit the benefits of DNNs \cite{zhang2017dncnn,santhanam2017rbdn,zhang2018ffdnet} and Transformer \cite{liu2021swin,liang2021swinir,zhang2022scunet}.
In addition, many image denoising models \cite{plotz2017benchmarking,brooks2019unprocessing} train on real image pairs captured by one camera.
However, these methods often have poor performance on other cameras.
More recently, denoising methods aim to approach the real-world denoising problem by using more realistic noisy images in training. This is achieved by making use of datasets with real-noisy~(or approximately real noisy) images~\cite{anwar2019real, kim2020transfer}, by using camera pipelines to synthesize real noisy images~\cite{jaroensri2019generating, cao2021pseudo, zamir2020cycleisp, conde2022model}, or using generative methods to model the real noises~\cite{chen2018image, kim2019grdn, yue2020dual, chang2020learning, vo2021hi, jang2021c2n, maleky2022noise2noiseflow}. Some methods use self-supervised training on clean real images without explicitly modeling real noises~\cite{wu2020unpaired, zheng2021unsupervised, kim2021noise2score, xie2020noise2same, huang2021neighbor2neighbor, quan2020self2self}.
While image based denoising methods can in theory be used for videos by treating each frame as a separate image, directly doing so ignores the fruitful temporal connections between different frames in a video.

\vspace{0.5mm}
\noindent\textbf{Video denoising}
aims at removing noise to synthesize clean videos.
Based on BM3D \cite{dabov2007bm3d}, VBM4D \cite{maggioni2012bm4d} presents a video filtering algorithm to exploit temporal and spatial redundancy of a video.
Early works~\cite{chen2016drnn} use recurrent networks to capture sequential information.
Recently, BasicVSR++ \cite{chan2021basicvsrpp} improves the second-order grid propagation and flow-guided deformable alignment in RNN and extends video super-resolution to the video denoising \cite{chan2022basicvsrpp2}.
In addition, some denoising methods adopt an asymmetric loss function \cite{vogels2018kpnn} to optimize the networks, or propose patch-based video denoising algorithm \cite{arias2018vnlb,davy2018vnlnet} to exploit the correlations among patches.
PaCNet \cite{vaksman2021patch} combines a patch-based framework with CNN by augmenting video sequences with patch-craft frames.
To further improve over patch-based methods, DVDnet \cite{tassano2019dvdnet} proposes spatial and temporal denoising blocks and trains them separately.
To boost the efficiency, FastDVDnet \cite{tassano2020fastdvdnet} extends DVDnet \cite{tassano2019dvdnet} by using two denoising steps which composed of a modified multi-scale U-Net~\cite{ronneberger2015u}. 
A few works also propose to improve performance with unknown noises by using self-supervised learning~\cite{sheth2021unsupervised, dewil2021self, dewil2022self}.
Recently, ViDeNN \cite{claus2019videnn} proposes a blind denoising method trained either on AWGN noise or on collected real-world videos. VRT \cite{liang2022vrt} proposes a video restoration transformer with parallel frame prediction, and achieves the state-of-the-art performance in video denoising. 
Existing optical-flow-based methods~\cite{wang2019edvr,liang2022vrt,li2022flornn,chan2021basicvsrpp, yu2020joint} mostly did not consider the impact of noises on the flow estimation. However, it is known that most flow estimation networks tend to deteriorate under noise~\cite{xue2019toflow}. 
This can lead to wrong feature alignment. 
Although TOFlow \cite{xue2019toflow} learns the task-oriented flow, it neglects to further refine and exploit more compensation information when the flow is still not precise.
To complete the discussion on existing video denoising works, we summarize the different setups in Figure~\ref{fig:diff}.


\newpage
\section{Proposed Method}
In this paper, we aim to design a new architecture for synthesized Gaussian denoising and real video denoising to synthesize a clean video from a noisy video sequence $\{\bI_1, \ldots, \bI_T\}$.
The proposed architecture is provided in Figure \ref{fig:arch}.
Our proposed architecture is multi-scale and it contains a down-scaling and a up-scaling processing.
At each scale, the model has a denoising-oriented flow refinement and mutual feature alignment (MFA) with the flow-guided denoising propagation.
First, we use an encoder to extract low-level features of a given noisy video sequence. 
On the other hand, we propose to refine the optical flows between local noisy neighbors at each scale.
Second, we use the refined optical flows guilds the module for better feature alignment and propagation.
Last, we further remove the spatial noise in the up-scaling.
Note that all modules are differentiable and can be train in an end-to-end manner.

\subsection{Denoising-Oriented Flow Refinement}
The noise in each frame can be regarded as pixel occlusion.
Different levels of noise produce different occlusions which thus affect the quality of the estimated optical flow, especially for strong noise as shown in Figure \ref{fig:noise_flow}.
Under the high level of noise, using an inaccurate optical flow may lead to misleading feature alignment and propagation.
To address this, we propose a denoising-oriented flow refinement. 

Without using the optical flow of the original size, we calculate the optical flows of the down-scaling frames.
Specifically, we first downscale the noisy frames as $\bI^{\downarrow s}$ using a bicubic operation at $\frac{1}{s}$ resolution.
Then, the optical flow between adjacent downsampled frames $\bI^{\downarrow s}_{i}$ and $\bI^{\downarrow s}_{j}$ can be computed by a flow estimation network $F$, \ie
\begin{align}
    \hat{\bo}_{i{\to}j}^s = \mF \left(\bI^{\downarrow s}_{i}, \bI^{\downarrow s}_{j}\right),
\end{align}
where $\downarrow_s$ is downsampling with the scale $s$, and $\mF$ is a flow estimation network. 

A video contains a long-range of temporal information.
To exploit such information, the forward and backward optical flows are important for feature propagation \cite{chan2021basicvsrpp}.
To this end, we propose to minimize the following $\ell_1$ loss to refine the optical flows, \ie
\begin{align}
    \mL_{\text{flow}} = \sum\nolimits_{s, t} \left\| \hat{\bo}_{t{\to}t+\Delta t}^s - {\bo}_{t{\to}t+\Delta t}^s \right\|_1,
\end{align} 
where the scale $s{\in}\{2,4\}$, the neighbor $\Delta t{\in}\{{\pm}1\}$, and ${\bo}_{t{\to}t+\Delta t}^s$ are the pseudo ground-truth forward or backward flows of noise-free clean videos. Specifically, the flow network $\mF_{fixed}$ used to calculate ${\bo}_{t{\to}t+\Delta t}^s$ is fixed. The flow network $\mF$ used to calculate $\hat{\bo}_{t{\to}t+\Delta t}^s$ is learnable. In our experiment, both networks use the same architecture and are initialized by the pre-trained weights of SpyNet~\cite{ranjan2017optical}.
In the experiment, we train the flow estimation network in an end-to-end manner.
In this way, the optical flows are denoising-oriented and robust to different levels of noise.

\begin{figure*}[t]
  \begin{center}
  \vspace{-1mm}
  \includegraphics[width=1\linewidth]{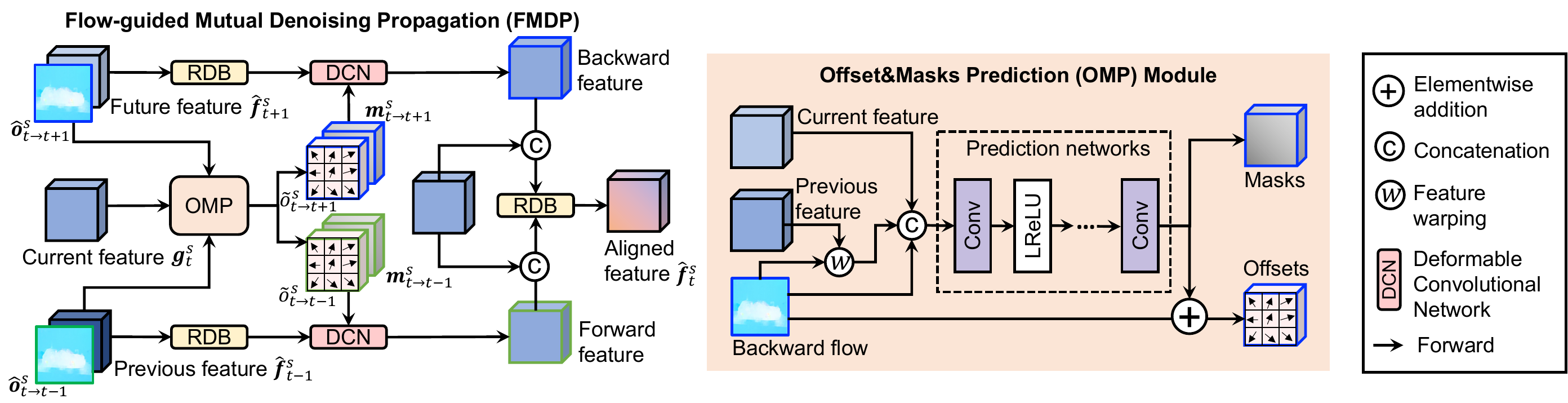}
  \end{center}
  \vspace{-3mm}
  \caption{The architecture of the flow-guided denoising mutual-propagation. Given previous and future features, refined optical flows, and current features, we first estimate the offsets and masks in DCN, and then we obtain the backward and forward features. Last, we fuse these features with a convolution to obtain the final aligned features.}
  \label{fig:propagation}
  \vspace{-4mm}
\end{figure*} 

\newpage
\subsection{Flow-guided Mutual Denoising Propagation} 
With the help of our denoising-oriented flow refinement, optical flows contain temporal information which is important for feature alignment and propagation.
Therefore, we propose a flow-guided denoising propagation module.
Specifically, given an $T$-frame noisy video $\{\bI_1, \ldots, \bI_T\}$, we first deploy a convolutional layer as the feature extractor to learn low-level features $\{\widehat{\bg}_1, \ldots, \widehat{\bg}_n\}$.
Then we use the residual block to extract deep features $\bg_t^s$ and reduce the noise at the $i$-th time step of the $s$-scale.
On the other hand, we use our denoising-oriented flow network to estimate the forward flows $\hat{\bo}_{t{\to}t{-}1}$ and backward flows $\hat{\bo}_{t{\to}t{+}1}$.
Last, based on the denoising features $\bg_i^s$, neighboring features $\bff_{t{-}1}^s$ and $\bff_{t{+}1}^s$ and the refined flows, we are able to mutually guide the denoising propagation to learn the denoising features $\hat{\bff}_t^s$ at the current time step of the $s$-scale.
Formally, we define a flow-guided denoising propagation as:
\begin{align}
    \!\!\!\hat{\bff}_t^s {=} \mA \left(\mP_f(\bg_t^{s}, \bff_{t{-}1}^s, \hat{\bo}_{t{\to}t{-}1}), \mP_b(\bg_t^{s}, \bff_{t{+}1}^s, \hat{\bo}_{t{\to}t{+}1}) \right)\!\!,\!\!\!\!
\end{align}
where $\mA$ is an aggregation function which is implemented as a $1{\times}1$ convolutional layer in the expriment, $\mP_f$ is the forward propagation function  and $\mP_b$ is the backward propagation function.
Our mutual propagation module is able to jointly propagate forward and backward features in the next propagation, which is different from BasicVSR++ \cite{chan2021basicvsrpp} which use backward and forward propagation in order in one scale.
In contrast, our forward and backward propagation are processed in multiple scales which can aggregate more information for denoising.

Next, we will model the procedure of the forward propagation, and the procedure of the backward propagation can be formulated similarly but in the opposite direction.
Given the denoising features $\bg_i^s$, neighboring features $\bff_{t{-}1}^s$ and the refined flows $\hat{\bo}_{t{\to}t{-}1}$, we define the forward propagation as:
\begin{align}
    \mP_f(\bg_t^{s}, \bff_{t{-}1}^s, \hat{\bo}_{t{\to}t{-}1}) = \mR \left( \bg_t^s, \mT(\bff_{t{-}1}^s, \hat{\bo}_{t{\to}t{-}1}) \right),
\end{align}
where $\mR$ consists of multiple residual layers, \ie RDB, and $\mT$ is the spatial texture transfer function according to the optical flows, which can be defined as:
\begin{align}
    \mT(\bff_{t{-}1}^s, \hat{\bo}_{t{\to}t{-}1}) = \mD(\bff_{t{-}1}^s, \widetilde{\bo}^s_{t\rightarrow t{-}1}, \bmm^s_{t\rightarrow t{-}1}),
\end{align}
where $\mD$ is a deformable convolutional network (DCN) \cite{zhu2019dcnv2}, $\widetilde{\bo}^s_{t\rightarrow t{-}1}$ and $\bmm^s_{i\rightarrow i-p}$ are the offsets and masks. 
The relationship between our refined optical flows and the learned offsets can promote each other. 
The refined optical flows initialize the meaningful sampling locations and use the these flows to learn a set of offsets. 
These offsets have large diversity and provide more flexible locations for sampling in DCN.
These sampling locations allow the model to discover more meaningful relevant texture in a local region and reduce warping error.
As a result, these diverse offsets relieve the effects of the noise occlusion.
On the other hand, these learned offsets provide positive feedback to further update the optical flows.
The offsets and masks are formulated as
\begin{align}
    \widetilde{\bo}^s_{t\rightarrow t-1} &= \hat{\bo}^s_{t \rightarrow t-1} + c_1 \left(\left[\bg_t^s; \bar{\bff}_{t-1}^s; \hat{\bo}^s_{t \rightarrow t-1} \right]\right), \\
    \bmm^s_{i\rightarrow t-1} &= \tau (c_2\left(\left[\bg_t^s; \bar{\bff}_{t-1}^s; \hat{\bo}^s_{t \rightarrow t-1} \right]\right)),
\end{align}
where $\tau$ is a Sigmoid function, $[\cdot;\cdot]$ is a concatenation operation, $c_1$ and $c_2$ are convolutional layers, and $\bff_{t-1}^s$ is a warped feature using the optical flow $\hat{\bo}^s_{t\rightarrow t-1}$, \ie 
\begin{align}
    \bar{\bff}_{t-1}^s = \omega (\bff_{t-1}^s, \hat{\bo}^s_{t\rightarrow t-1}),
\end{align}
where $\omega(\cdot)$ is a warp function according to the optical flow.
Based on the above formulations, we obtain the propagation features in each scale.
At the last scale, we apply RDB blocks and a convolution layer as the noise prediction to learn the final residual and use the skip connection to obtain the denoised videos.

In this prorogation module, we do not directly use the second-order flows since the noise affects the accuracy of computation of flows and such error will be accumulated in the next propagation when the first-order optical flow is inaccurate, as shown in the supplementary. 
Our method can be extended to use second-order flows which can be our flow refinement module to reduce the propagation error.

\begin{figure*}[t]
    \begin{center}
    \vspace{-0.3cm}
    \includegraphics[width=0.9\linewidth]{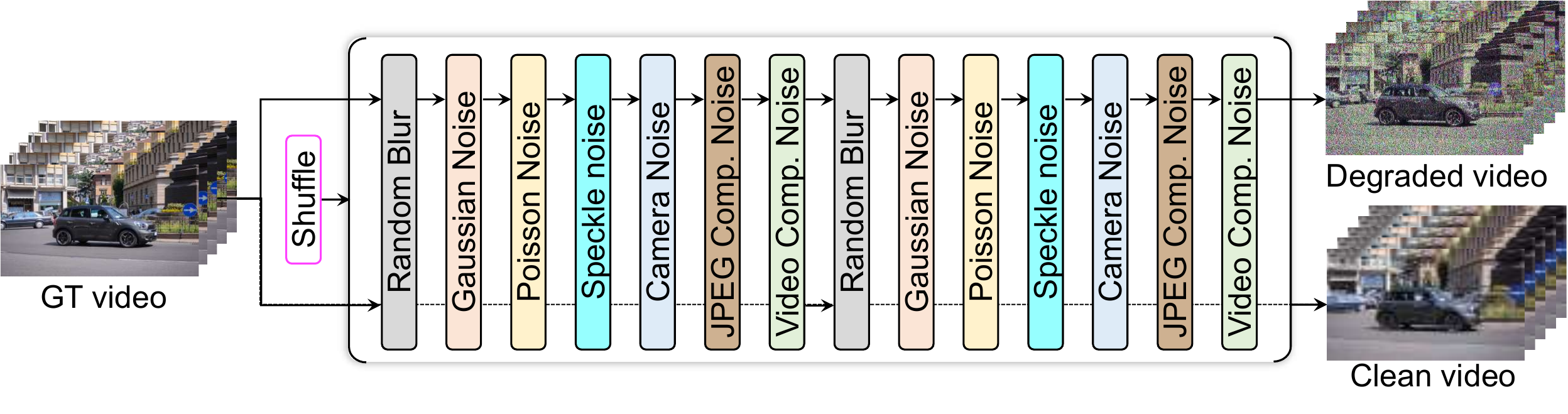}
    \end{center}
    \vspace{-0.6cm}
    \caption{An illustration of the proposed noise degradation pipeline. 
    For a high-quality video, a randomly shuffled degradation sequence is performed to produce a noisy video. The dash line means that we do not pass a certain degradation.}
    \label{fig:pipeline}
    \vspace{-0.4cm}
\end{figure*}

\newpage
\subsection{Real Noise Degradations}
Real noise distribution is different from Gaussian noise, which is purely additive and signal-dependent.
Training on Gaussian noise may have poor performance on real noise distribution.
Real-world videos often contain unknown noises and blur, and they differ from video to video. 
To cover the real noise distribution, we propose to a large range of noise, including signal dependent and independent noise.
Formally, given a GT video $\bI_{{HR}}$, we use the following shuffled degradation process to synthesize a noisy and a clean video: 
\begin{align}
    (\bI, \bI_{{clean}}) = \left( g_{i_1} \circ g_{i_2} \circ \cdots \circ g_{i_N} \right) (\bI_{{HR}}), 
\end{align}
where $\{i_1, \ldots, i_N\} = \phi(\{1, \ldots, N\})$, $N$ is the number of degradations, $\phi$ is a shuffle function, $\circ$ is a function composition, and $g_{i_n}$ is a degradation model of the $i_n$-th type. 

In practice, we propose use the following noise degradations, including
Gaussian noise, Poisson noise, Speckle noise, Processed camera sensor noise, JPEG compression noise and video compression noise.
We use Gaussian and Poisson noise because the noise in raw domain contains the read noise and shot noise which can be modeled by Gaussian and Poisson noise \cite{foi2008practical,mildenhall2018burst}.
Speckle noise exists in the synthetic aperture radar (SAR), medical ultrasound and optical coherence tomography images.
Besides, we also consider processed camera sensor noise which originates from the image signal processing (ISP). Inspired by \cite{zhang2022scunet}, the reverse ISP pipeline first get the raw image from an RGB image, then the forward pipeline constructs noisy raw image by adding noise to the raw image.
For the digital images storage problem, JPEG compression noise and video compression noise are very important.
Existing noise degradation models ignore the video compression noise, which are different from our model.
More details of above degradations are put in the supplementary materials.

Apart from noise, most real videos inherently suffer from blurriness.
Thus, we additionally consider two common blur degradations, \ie Gaussian blur and resizing blur.
As suggested in \cite{zhang2022scunet}, we apply the same blur degradations on both noisy video and its clean counterpart, which is very different from existing super-resolution degradation models \cite{zhang2021designing,wang2021realesrgan}.  
The reason is that blur degradations will change the resolution of latent clean videos of the noisy videos.

\begin{table*}[t]
  \vspace{-1mm}
  \caption{Quantitative comparison (average RGB channel PSNR/SSIM) with state-of-the-art methods for video denoising on the DAVIS and Set8 datasets. The best results are in $\textbf{bold}$.}
  \label{tab:comp_AWGN}
  \centering
  \resizebox{0.99\textwidth}{!}{
  \begin{tabular}{|c|c||cccccccc|c|}
    \hline
    \multicolumn{1}{|l|}{Datasets} & $\sigma$ & VBM4D \cite{maggioni2012bm4d} & VNLB \cite{arias2018vnlb} & DVDnet \cite{tassano2019dvdnet} & FastDVDnet \cite{tassano2020fastdvdnet} & VNLNet \cite{davy2018vnlnet} & PaCNet \cite{vaksman2021pacnet} & BasicVSR++ \cite{chan2022basicvsrpp2} & VRT \cite{liang2022vrt} & \bf{Ours} \\
    \hline\hline
    \multirow{6}{*}{\makecell{\scalebox{1}{DAVIS}\\\cite{khoreva2018davis}} } 
    & 10 & 37.58/- & 38.85/- & 38.13/.9657 & 38.71/.9672 & 39.56/.9707 & 39.97/.9713 & 40.13/.9754 & 40.82/.9776 & \bf{41.11}/\bf{.9797} \\
    & 20 & 33.88/- & 35.68/- & 35.70/.9422 & 35.77/.9405 & 36.53/.9464 & 37.10/.9470 & 37.41/.9598 & 38.15/.9625 & \bf{38.61}/\bf{.9677} \\
    & 30 & 31.65/- & 33.73/- & 34.08/.9188 & 34.04/.9167 & -/-    & 35.07/.9211 & 35.74/.9456 & 36.52/.9483 & \bf{37.10}/\bf{.9569} \\
    & 40 & 30.05/- & 32.32/- & 32.86/.8962 & 32.82/.8949 & 33.32/.8996 & 33.57/.8969 & 34.49/.9319 & 35.32/.9345 & \bf{35.98}/\bf{.9465} \\
    & 50 & 28.80/- & 31.13/- & 31.85/.8745 & 31.86/.8747 & -/-     & 32.39/.8743 & 33.45/.9179 & 34.36/.9211 & \bf{35.08}/\bf{.9363} \\
    & \!\!Avg.\!\! & 32.39/- & 34.34/- & 34.52/.9195 & 34.64/.9188 & -/- & 35.62/.9221 & 36.24/.9461 & 37.03/.9488 & \bf{37.58}/\bf{.9574} \\
    \hline
    \multirow{6}{*}{\makecell{\scalebox{1}{Set8}\\\cite{tassano2019dvdnet}} }
    & 10 & 36.05/- & 37.26/- & 36.08/.9510 & 36.44/.9540 & 37.28/.9606 & 37.06/.9590 & 36.83/.9574 & 37.88/.9630 & \bf{38.07}/\bf{.9661} \\
    & 20 & 32.18/- & 33.72/- & 33.49/.9182 & 33.43/.9196 & 34.02/.9273 & 33.94/.9247 & 34.15/.9319 & 35.02/.9373 & \bf{35.41}/\bf{.9456} \\
    & 30 & 30.00/- & 31.74/- & 31.68/.8862 & 31.68/.8889 & -/-     & 32.05/.8921 & 32.57/.9095 & 33.35/.9141 & \bf{33.87}/\bf{.9272} \\
    & 40 & 28.48/- & 30.39/- & 30.46/.8564 & 30.46/.8608 & 30.72/.8622 & 30.70/.8623 & 31.42/.8889 & 32.15/.8928 & \bf{32.76}/\bf{.9101} \\
    & 50 & 27.33/- & 29.24/- & 29.53/.8289 & 29.53/.8351 & -/-     & 29.66/.8349 & 30.49/.8690 & 31.22/.8733 & \bf{31.88}/\bf{.8942} \\
    & \!\!Avg.\!\! & 30.81/- & 32.47/- & 32.29/.8881 & 32.31/.8917 & -/- & 32.68/.8946 & 33.09/.9113 & 33.92/.9160 & \bf{34.40}/\bf{.9286} \\
    \hline
  \end{tabular}  
  }
  \vspace{-2mm}
\end{table*}

\begin{figure*}[t!]
	\centering
	\renewcommand{\g}{-0.7mm}
	\renewcommand{\tabcolsep}{1.8pt}
	\renewcommand{\arraystretch}{1}
	\resizebox{0.98\linewidth}{!} {
	\hspace{-4.5mm}
	\begin{tabular}{cc}
			\renewcommand{\name}{figures/davis/00000019_}
			\renewcommand{\h}{0.12}
			\renewcommand{\w}{0.2}
			\begin{tabular}{cc}
				\begin{adjustbox}{valign=t}
					\begin{tabular}{c}
		         	\includegraphics[trim={0 0 0 0},clip, width=0.354\textwidth]{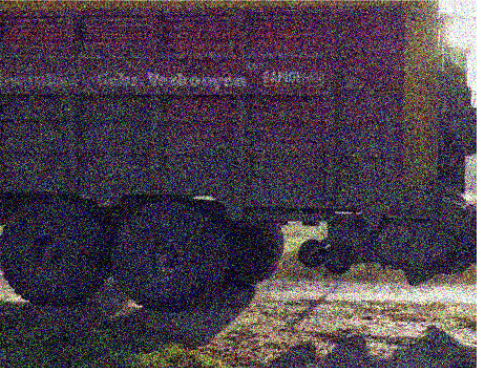}
						\\
						 00000019, tractor, DAVIS \cite{khoreva2018davis}
					\end{tabular}
				\end{adjustbox}
				\begin{adjustbox}{valign=t}
					\begin{tabular}{cccccc}
						\includegraphics[trim={0 0 0 0},clip,height=\h \textwidth, width=\w \textwidth]{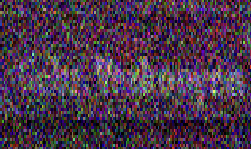} \hspace{\g} &
						\includegraphics[trim={0 0 0 0},clip,height=\h \textwidth, width=\w \textwidth]{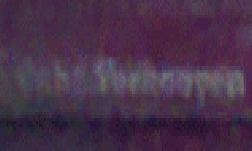} \hspace{\g} &
						\includegraphics[trim={0 0 0 0},clip,height=\h \textwidth, width=\w \textwidth]{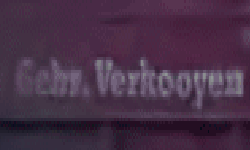} & 
						\includegraphics[trim={0 0 0 0},clip,height=\h \textwidth, width=\w \textwidth]{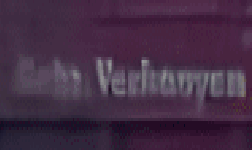} \hspace{\g} 
						\\
						Noisy frame \vspace{-0.5pt}($\sigma{=}50$)\vspace{-4.5pt} &
						VBM4D \cite{maggioni2012bm4d} & DVDnet \cite{tassano2019dvdnet} & FastDVDnet \cite{tassano2020fastdvdnet}
						\\
						\vspace{-2mm}
						\\
						\includegraphics[trim={0 0 0 0},clip,height=\h \textwidth, width=\w \textwidth]{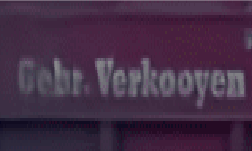} \hspace{\g} &
						\includegraphics[trim={0 0 0 0},clip,height=\h \textwidth, width=\w \textwidth]{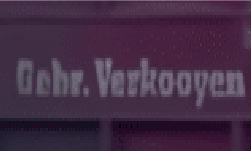} \hspace{\g} &
						\includegraphics[trim={0 0 0 0},clip,height=\h \textwidth, width=\w \textwidth]{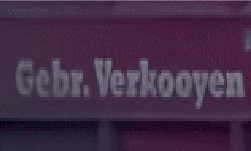}
						\hspace{\g} &		
						\includegraphics[trim={0 0 0 0},clip,height=\h \textwidth, width=\w \textwidth]{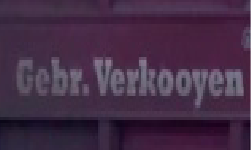}   
						\\ 
						BasicVSR++ \cite{chan2021basicvsrpp} \hspace{\g} &	VRT \cite{liang2022vrt} \hspace{\g} &
						\textbf{Ours} &
						GT
						\\
					\end{tabular}
				\end{adjustbox}
			\end{tabular}
			
		\end{tabular}
	}
    \resizebox{0.98\linewidth}{!} {
	\hspace{-4.5mm}
	\begin{tabular}{cc}
			\renewcommand{\name}{figures/davis/00000042_}
			\renewcommand{\h}{0.12}
			\renewcommand{\w}{0.2}
			\begin{tabular}{cc}
				\begin{adjustbox}{valign=t}
					\begin{tabular}{c}
		         	\includegraphics[trim={0 0 0 0},clip, width=0.354\textwidth]{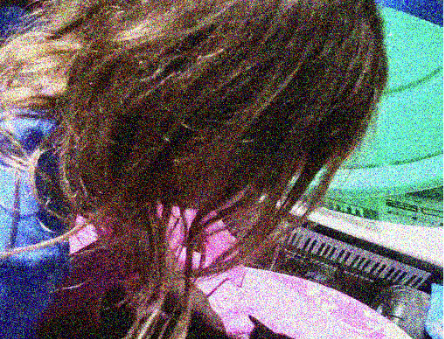}
						\\
						 00000042, cats-car, DAVIS \cite{khoreva2018davis}
					\end{tabular}
				\end{adjustbox}
				\begin{adjustbox}{valign=t}
					\begin{tabular}{cccccc}
						\includegraphics[trim={0 0 0 0},clip,height=\h \textwidth, width=\w \textwidth]{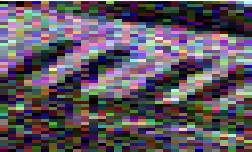} \hspace{\g} &
						\includegraphics[trim={0 0 0 0},clip,height=\h \textwidth, width=\w \textwidth]{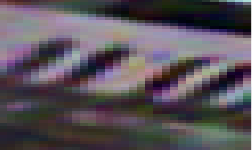} \hspace{\g} &
						\includegraphics[trim={0 0 0 0},clip,height=\h \textwidth, width=\w \textwidth]{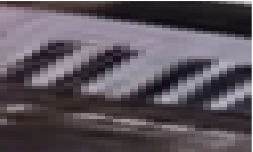} & 
						\includegraphics[trim={0 0 0 0},clip,height=\h \textwidth, width=\w \textwidth]{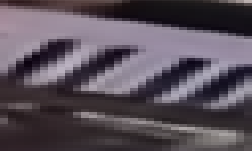} \hspace{\g} 
						\\
						\vspace{-0.5pt}Noisy frame ($\sigma{=}50$)\vspace{-4.5pt} &
						VBM4D \cite{maggioni2012bm4d} & DVDnet \cite{tassano2019dvdnet} & FastDVDnet \cite{tassano2020fastdvdnet}
						\\
						\vspace{-2mm}
						\\
						\includegraphics[trim={0 0 0 0},clip,height=\h \textwidth, width=\w \textwidth]{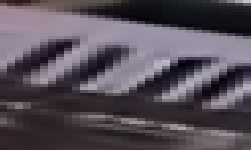} \hspace{\g} &
						\includegraphics[trim={0 0 0 0},clip,height=\h \textwidth, width=\w \textwidth]{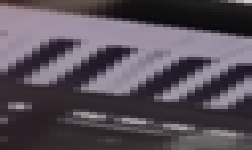} \hspace{\g} &
						\includegraphics[trim={0 0 0 0},clip,height=\h \textwidth, width=\w \textwidth]{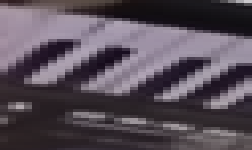}
						\hspace{\g} &		
						\includegraphics[trim={0 0 0 0},clip,height=\h \textwidth, width=\w \textwidth]{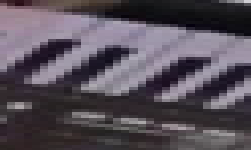}   
						\\ 
						BasicVSR++ \cite{chan2021basicvsrpp} \hspace{\g} &	VRT \cite{liang2022vrt} \hspace{\g} &
						\textbf{Ours} &
						GT
						\\
					\end{tabular}
				\end{adjustbox}
			\end{tabular}
		\end{tabular}
	}
	\caption{Visual comparison of different methods on DAVIS under the noise level of 50. More results are in Supplementary.} 
	\vspace{-5mm}
	\label{fig:sr_quali}
\end{figure*}

\newpage
\section{Experiments}

\subsection{Synthetic Gaussian Denoising}
\noindent\textbf{Datasets.}
We use DAVIS \cite{khoreva2018davis} and Set8 \cite{tassano2019dvdnet} in our synthetic Gaussian denoising experiments. 
We train all models on DAVIS training set \cite{khoreva2018davis}, and test them on DAVIS testing set and Set8 \cite{tassano2019dvdnet}.
Specifically, DAVIS \cite{khoreva2018davis} contains 90 training sequences (6208 frames) at 480p resolution, and 30 testing sequences (2086 frames) of resolution 854${\times}$480.
Set8 \cite{tassano2019dvdnet} has 8 testing video sequences with a resolution of 960${\times}$540.
Following the setting of \cite{liang2022vrt}, we synthesize the noisy video sequences by adding AWGN with noise level $\sigma{\in}[0, 50]$ on the DAVIS \cite{khoreva2018davis} training set.
We then train the model by using the synthesized data and test it on the DAVIS testing set and Set8 \cite{tassano2019dvdnet} with different Gaussian noise levels $\{10, 20, 30, 40, 50\}$.
We compare our model with the following state-of-the-art video denoising methods, including VBM4D \cite{maggioni2012bm4d}, VNLB \cite{arias2018vnlb}, DVDnet \cite{tassano2019dvdnet}, FastDVDnet \cite{tassano2020fastdvdnet}, VNLNet \cite{davy2018vnlnet}, PaCNet \cite{vaksman2021pacnet}, BasicVSR++ \cite{chan2022basicvsrpp2}, and VRT \cite{liang2022vrt}.

\vspace{1mm}
\noindent\textbf{Quantitative comparison.}
In Table \ref{tab:comp_AWGN}, we show PSNR and SSIM of different methods on the DAVIS testing set \cite{khoreva2018davis} and Set8 \cite{tassano2019dvdnet} under different noise levels.
Compared with other methods, our method has the best performance on both DAVIS and Set8 with a large margin.
Specifically, our model outperforms BasicVSR++ \cite{chan2022basicvsrpp2} and previous SOTA VRT \cite{liang2022vrt} by an average PSNR of \textbf{1.34dB} and \textbf{0.55dB}, respectively.
These methods are influenced by noise since they neglect to learn robust flow networks.
Moreover, the improvement of our method increases as the noise levels increase with the help of denoising-oriented pyramid flows.
These results demonstrate the superiority of our proposed architecture.

\vspace{1mm}
\noindent\textbf{Qualitative comparison.}
In Figure \ref{fig:sr_quali}, we compare different methods under the high noise level of 50.
Our proposed denoiser restores better structures and preserves clean edge than previous state-of-the-art video denoising methods, even though the noise level is high. 
In particular, our model is able to restore the letters `Gebr' in the first example and piano texture in the second example of Figure \ref{fig:sr_quali}. %
In contrast, VBM4D \cite{maggioni2012bm4d}, DVDnet \cite{tassano2019dvdnet} and FastDVDnet \cite{tassano2020fastdvdnet} fail to remove severe noise from a video frame.
BasicVSR++ \cite{chan2021basicvsrpp} and VRT \cite{liang2022vrt} only restore part of the textures.

\begin{table*}[t]
  \vspace{-0.2cm}
  \caption{Quantitative comparison (average RGB channel PSNR) with state-of-the-art methods for video denoising on the CRVD (indoor) datasets \cite{yue2020crvd}. The best results are in $\textbf{bold}$. The superscript $^*$ indicates the results are from the paper of RViDeNet \cite{yue2020rvidenet}. }
  \label{tab:comp_crvd}
  \centering
  \resizebox{1\textwidth}{!}{
  \begin{tabular}{|c||cccccccc|c|}
    \hline
    ISO & ViDeNN$^*$ \cite{claus2019videnn} & VBM4D$^*$ \cite{maggioni2012bm4d} & TOFlow$^*$ \cite{xue2019toflow} & SMD$^*$ \cite{chen2019smd} & EDVR$^*$ \cite{wang2019edvr} & DIDN$^*$ \cite{yu2019didn} & RViDeNet$^*$ \cite{yue2020rvidenet} & FloRNN \cite{li2022flornn} &\bf{Ours} \\
    \hline\hline
    1600  & 35.44/0.966 & 39.34/0.967 & 37.61/0.964 & 37.81/0.969 & 42.10/0.984 & 41.85/0.985 & 43.13/0.988 & 44.07/0.991 & \bf{44.10}/{0.992} \\
    3200  & 34.37/0.946 & 36.62/0.951 & 36.97/0.958 & 37.07/0.964 & 41.03/0.980 & 40.65/0.980 & 41.99/0.985 & 42.98/0.988 & \bf{43.38}/{0.990} \\
    6400  & 31.87/0.880 & 33.75/0.925 & 35.42/0.940 & 35.93/0.958 & 38.98/0.974 & 38.82/0.975 & 39.99/0.980 & 41.04/0.985 & \bf{41.15}/{0.986} \\
    12800 & 29.79/0.778 & 31.59/0.902 & 33.54/0.910 & 34.91/0.952 & 37.47/0.967 & 37.54/0.970 & 38.44/0.975 & 39.64/0.981 & \bf{39.89}/{0.982} \\
    25600 & 25.95/0.559 & 29.48/0.868 & 30.52/0.835 & 33.64/0.942 & 35.26/0.957 & 35.28/0.960 & 36.21/0.968 & 37.34/0.976 & \bf{37.56}/{0.979} \\
    \hline
    Avg.  & 34.16/0.922 & 34.81/0.921 & 34.81/0.921 & 35.87/0.957 & 38.97/0.972 & 38.83/0.974 & 39.95/0.979 & 41.01/0.984 & \bf{41.22}/{0.986} \\ 
    \hline
  \end{tabular}
  }\vspace{-2mm}
\end{table*}

\begin{figure*}[t]
	\centering
	\renewcommand{\g}{-0.7mm}
	\renewcommand{\tabcolsep}{1.8pt}
	\renewcommand{\arraystretch}{1}
	\resizebox{1\linewidth}{!} {
	\hspace{-4.5mm}
	\begin{tabular}{cc}
			\renewcommand{\name}{figures/crvd/ISO25600frame4}
			\renewcommand{\h}{0.12}
			\renewcommand{\w}{0.2}
			\begin{tabular}{cc}
				\begin{adjustbox}{valign=t}
					\begin{tabular}{c}
		         	\includegraphics[trim={20 0 0 0},clip, width=0.354\textwidth]{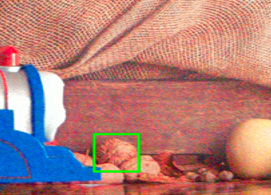}
						\\
						 scene10, ISO25600, frame4 
					\end{tabular}
				\end{adjustbox}
				\begin{adjustbox}{valign=t}
					\begin{tabular}{cccccc}
						\includegraphics[trim={0 0 0 0},clip,height=\h \textwidth, width=\w \textwidth]{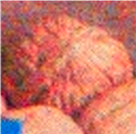} \hspace{\g} &
						\includegraphics[trim={0 0 0 0},clip,height=\h \textwidth, width=\w \textwidth]{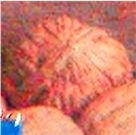} \hspace{\g} &
						\includegraphics[trim={0 0 0 0},clip,height=\h \textwidth, width=\w \textwidth]{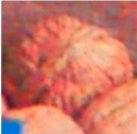} & 
						\includegraphics[trim={0 0 0 0},clip,height=\h \textwidth, width=\w \textwidth]{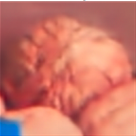} \hspace{\g} 
						\\
						Noisy frame (26.83dB) \vspace{-0.5pt}\vspace{-4.5pt} &
						ViDeNN \cite{claus2019videnn} (28.25dB) & TOFlow \cite{xue2019toflow} (31.76dB) & SMD \cite{chen2019smd} (31.31dB)
						\\
						\vspace{-2mm}
						\\
						\includegraphics[trim={0 0 0 0},clip,height=\h \textwidth, width=\w \textwidth]{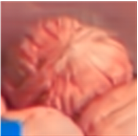} \hspace{\g} &
						\includegraphics[trim={0 0 0 0},clip,height=\h \textwidth, width=\w \textwidth]{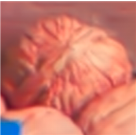} \hspace{\g} &
						\includegraphics[trim={0 0 0 0},clip,height=\h \textwidth, width=\w \textwidth]{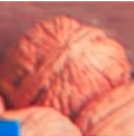}
						\hspace{\g} &		
						\includegraphics[trim={0 0 0 0},clip,height=\h \textwidth, width=\w \textwidth]{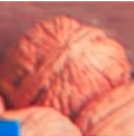}   
						\\ 
						DIDN \cite{yu2019didn} (33.82dB) \hspace{\g} &	RViDeNet \cite{yue2020rvidenet} (35.88dB) \hspace{\g} &
						\textbf{Ours} (\textbf{36.90dB}) &
						GT (PSNR)
						\\
					\end{tabular}
				\end{adjustbox}
			\end{tabular}
			
		\end{tabular}
	}
	\vspace{-1mm}
	\caption{Visual comparison and PSNR (dB) of different methods on CRVD indoor dataset \cite{yue2020crvd}.} 
	\vspace{-4mm}
	\label{fig:sr_quali_crvd}
\end{figure*}

\subsection{Real-world Video Denoising}


\vspace{-2mm}
\subsubsection{Results on the CRVD Dataset}
\vspace{-2mm}
The CRVD dataset \cite{yue2020crvd} is captured in the raw domain and it contains 6 indoor scenes for training and 5 indoor scenes for testing. Besides, it has 10 dynamic outdoor scenes without ground-truth clean videos. For each scene, the dataset has 5 different ISO settings. 
To evaluate our model on this dataset, we first apply a trained ISP model to raw data to generate sRGB videos, then we train our model on 6 indoor scenes.

In Table \ref{tab:comp_crvd}, we mainly compare raw video denoising methods in the sRGB domain under different ISO settings.
Our method has the highest PSNR and SSIM across all ISO settings and thus achieves the best performance.
In Figure \ref{fig:sr_quali_crvd}, our method has higher PSRN with more high-frequency details and less noise.
In contrast, ViDeNN \cite{claus2019videnn} and TOFlow \cite{xue2019toflow} still retain most of the noise in the video frames.
Other methods remove the high-frequency texture when denoising.
In addition, because the CRVD outdoor videos do not have ground-truth clean videos, we compare the visual results of different methods in the supplementary materials.

\begin{figure}[t]
  \begin{center}
  \includegraphics[width=1\linewidth]{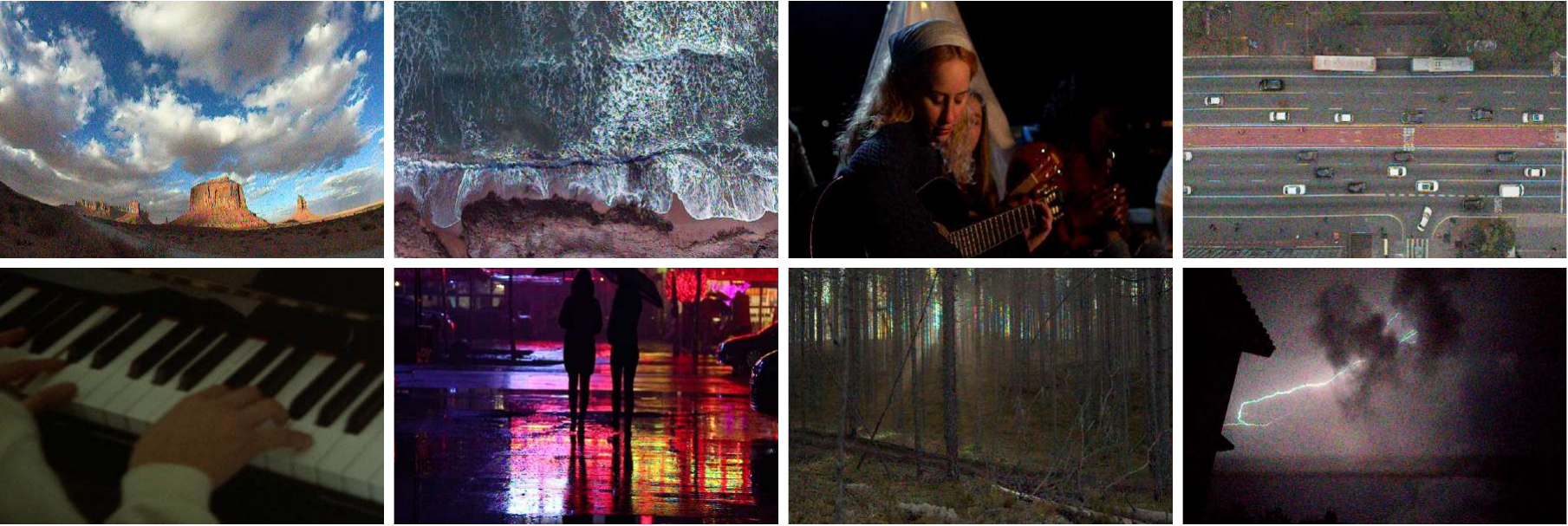}
  \end{center}
  \vspace{-3mm}
  \caption{Our NoisyVideo consists of videos with a wide range of content and different types of unknown noise. It can be served as a common ground for future comparison of real video denoising.}
  \label{fig:data_exp}
  \vspace{-3mm}
\end{figure}

\begin{table}[t]
  \caption{Quantitative Comparison of different methods on our RealNoise dataset for the real video denoising task. 
  }
  \label{tab:real_comp}
  \centering
  \resizebox{0.42\textwidth}{!}{
  \begin{tabular}{|l||ccc|}
    \hline
    Methods & NIQE$\downarrow$ & BRISQUE$\downarrow$ & PIQE$\downarrow$  \\
    \hline\hline
    VBM4D \cite{maggioni2012bm4d}             & 4.2692 & 43.4138 & 56.7153 \\
    Restormer \cite{zamir2021restormer}       & 3.7348 & 34.2775 & 47.8616 \\
    SCUNet \cite{zhang2022scunet}             & 3.3543 & 31.1964 & 42.3837  \\
    ViDeNN  \cite{claus2019videnn}            & 4.2911 & 21.9146 & 30.5148 \\
    FloRNN \cite{li2022flornn}                & 4.2568 & 33.4969 & 55.9694 \\
    BasicVSR++ \cite{chan2021basicvsrpp}~~~~~~~~  & 3.7767 & 24.7670 & 29.3206 \\
    \hline
    Ours-real  & 3.2818 & 21.7689 & 30.2147 \\
    \hline
	\end{tabular}
  }
  \vspace{-6mm}
\end{table}

\vspace{-3mm}
\subsubsection{Results on the Proposed RealNoise Video Dataset}
\vspace{-2mm}
To evaluate performance on more realistic videos, we additionally propose a new benchmark test dataset for real-world video denoising, called RealNoise.
This dataset is collected from Pexels, with videos under the Creative Common license. 
It contains of 10 diverse scene types with different unknown noises, and each video contains at least 100 frames with no scene change.
Examples of the dataset are shown in Figure~\ref{fig:data_exp} and further details are provided in the supplementary.

For training of the real video denoising model, we use REDS~\cite{nah2019reds} as the training set.
To have a fair comparison with other methods, we follow the setting of~\cite{wang2019edvr}, and use 266 regrouped training clips and 4 testing clips (denoted as REDS4), where each with 100 consecutive frames.
During training, we synthesize noisy videos on the REDS training set by using our proposed noise degradation model.

\begin{figure*}[t]
	\centering
	\renewcommand{\g}{-0.7mm}
	\renewcommand{\tabcolsep}{1.8pt}
	\renewcommand{\arraystretch}{1}
    \resizebox{1\linewidth}{!} {
	\hspace{-4.5mm}
	\begin{tabular}{cc}
			\renewcommand{\name}{figures/ourdata/00000003_}
			\renewcommand{\h}{0.12}
			\renewcommand{\w}{0.2}
			\begin{tabular}{cc}
				\begin{adjustbox}{valign=t}
					\begin{tabular}{c}
		         	\includegraphics[trim={0 0 0 0},clip, width=0.354\textwidth]{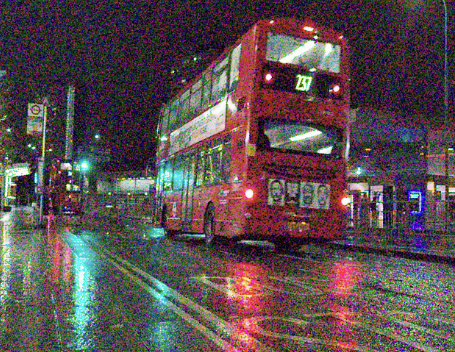}
						\\
						 RealNoise, street, frame3
					\end{tabular}
				\end{adjustbox}
				\begin{adjustbox}{valign=t}
					\begin{tabular}{cccccc}
						\includegraphics[trim={0 0 0 0},clip,height=\h \textwidth, width=\w \textwidth]{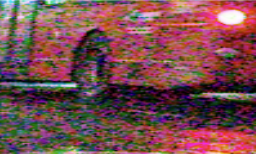} \hspace{\g} &
						\includegraphics[trim={0 0 0 0},clip,height=\h \textwidth, width=\w \textwidth]{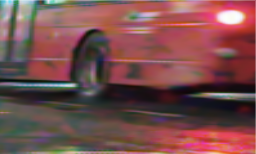} \hspace{\g} &
						\includegraphics[trim={0 0 0 0},clip,height=\h \textwidth, width=\w \textwidth]{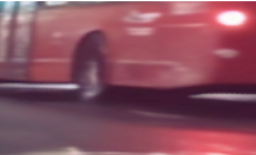} & 
						\includegraphics[trim={0 0 0 0},clip,height=\h \textwidth, width=\w \textwidth]{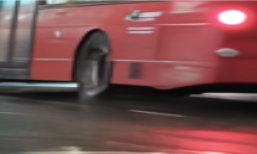} \hspace{\g} 
						\\
						Noisy frame \vspace{-0.5pt}\vspace{-4.5pt} &
						VBM4D \cite{maggioni2012bm4d} & Restormer \cite{zamir2021restormer} & SCUNet \cite{zhang2022scunet}
						\\
						\vspace{-2mm}
						\\
						\includegraphics[trim={0 0 0 0},clip,height=\h \textwidth, width=\w \textwidth]{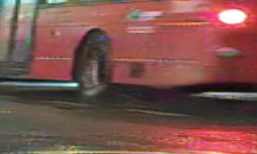} \hspace{\g} &
						\includegraphics[trim={0 0 0 0},clip,height=\h \textwidth, width=\w \textwidth]{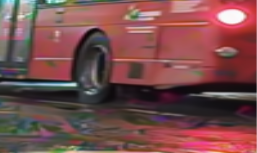} \hspace{\g} &
						\includegraphics[trim={0 0 0 0},clip,height=\h \textwidth, width=\w \textwidth]{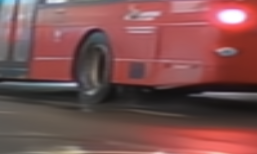}
						\hspace{\g} &		
						\includegraphics[trim={0 0 0 0},clip,height=\h \textwidth, width=\w \textwidth]{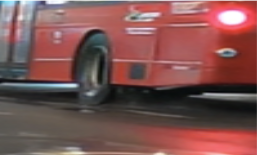}   
						\\ 
						ViDeNN \cite{claus2019videnn} \vspace{-0.5pt} \vspace{-4.5pt} & FloRNN \cite{li2022flornn}  &
						BasicVSR++ \cite{chan2022basicvsrpp2} & Ours \vspace{2mm}
						\\
					\end{tabular}
				\end{adjustbox}
			\end{tabular}
			
		\end{tabular}
	}
	\caption{Visual comparison of different methods on our proposed RealNoise dataset.} 
	\label{fig:realnoise_quali}
\end{figure*}

As shown in Figure \ref{fig:realnoise_quali}, the proposed method achchives better visual quality than other image restoration methods and camera-based denoising methods. Our method is able to synthesize sharper texture.  In contrast, traditional method VBM4D \cite{maggioni2012bm4d} only remove the part of noise.
Restormer \cite{zamir2021restormer} and SCUNet \cite{zhang2022scunet} are image denoising methods and have limited performance when the real noise is complex because they do not exploit the temporal information.
ViDeNN \cite{claus2019videnn} and FloRNN \cite{li2022flornn} introduce artifacts in the video since they are trained on a specific camera.
In Table \ref{tab:real_comp}, we compare different methods on our dataset. Here, we use three non-reference metrics NIQE \cite{mittal2012niqe}, BRISQUE \cite{mittal2011brisque} and PIQE \cite{venkatanath2015piqe} as evaluation metrics because they are commonly used to measure the quality of images and ground-truth videos are not available. Our method shows very competitive results in all three scores as well. Note that these metrics do not always match perceptual visual quality~\cite{lugmayr2020ntire}.
From the visual example, with the help of our noise degradation model, our denoiser is able to reduce real-world noise.

\begin{table*}[t]
    \begin{minipage}{.42\linewidth}
    \centering
    \caption{Abaltion study on DAVIS \cite{khoreva2018davis} ($\sigma{=}50$).}
    \label{tab:ablation}
    \resizebox{1\textwidth}{!}{
    \begin{tabular}{|c|ccccc|}
    \hline
    Multiscale & & \checkmark & \checkmark & \checkmark & \checkmark  \\
    FGDP  & & & \checkmark & \checkmark & \checkmark\\
    MFA   & & & & \checkmark & \checkmark \\
    Refine flow   & & & & & \checkmark\\
    \hline
    PSNR  &  33.46  & 34.40 & 34.67  & 34.90 & \bf{35.08} \\
    SSIM  &  0.9179 & 0.9220 & 0.9233 & 0.9278 & \bf{0.9363} \\
    \hline
	\end{tabular}
	}
  \end{minipage}
  \begin{minipage}{.32\linewidth}
    \centering
    \caption{PSNR for Clipped Gaussian noise.}
    \label{tab:clip}
    \resizebox{0.84\textwidth}{!}{
    \begin{tabular}{|l||ccc|}
    \hline
    & \multicolumn{3}{c|}{Noise levels (DAVIS)}   \\ \cline{2-4}
    \multicolumn{1}{|l||}{\multirow{-2}{*}{Methods}} & 10 & 30 & 50 \\
    \hline\hline
    ViDeNN \cite{claus2019videnn}     & 37.13 & 32.24 & 29.77 \\
    FastDVDnet \cite{tassano2020fastdvdnet} \!\! & 38.65 & 33.59 & 31.28 \\
    PaCNet   \cite{vaksman2021pacnet}    & 39.96 & 34.66 & 32.00  \\
    \hline
    Ours-blind & {40.94} & {36.79} & {34.65}  \\
    \textbf{Ours} & \bf{41.00} & \bf{36.91} & \bf{34.83}  \\
    \hline
	\end{tabular}
	}
  \end{minipage}%
  \begin{minipage}{.26\linewidth}
    \centering
    \caption{PSNR for image denoising.}
    \label{tab:comp_img}
    \resizebox{0.98\textwidth}{!}{
    \begin{tabular}{|l||ccc|}
    \hline
    & \multicolumn{3}{c|}{Noise levels (Set8)} \\ \cline{2-4}
    \multicolumn{1}{|l||}{\multirow{-2}{*}{Methods}}  & 15 & 25 & 50 \\
    \hline\hline
    BM3D \cite{dabov2007bm3d}   & 29.00 & 28.64 & 26.50 \\
    Restormer \cite{zamir2021restormer} & 34.36 & 31.40 & 28.57 \\
    SwinIR \cite{liang2021swinir} & 34.87 & 32.37 & 29.19 \\
    SCUNet \cite{zhang2022scunet} & 34.82 & 32.34 & 29.14 \\
    \hline
    \textbf{Ours} & \bf{36.47} & \bf{34.49} & \bf{31.77} \\
    \hline
	\end{tabular}
	}
  \end{minipage}
  \vspace{-3mm}
\end{table*}

\begin{table*}[h!]
    \begin{minipage}{.46\linewidth}
    \centering
    \caption{Ablation study on noise types on synthesized REDS4.}
    \label{tab:ablation_type}
    \resizebox{1\textwidth}{!}{
    \begin{tabular}{|l||cc|}
    \hline
    \multicolumn{1}{|l||}{Degradation types} & ~PSNR~ & ~SSIM~          \\ 
    \hline\hline
    w/o blur degradation                & 26.94 & 0.7783 \\  
    w/o processed camera sensor noise ~~~~~~~~  & 27.10 & 0.7799 \\  
    w/o video compression noise         & 27.02 & 0.7791 \\ 
    \hline
    w/ all degradations                 & 27.46 & 0.7912 \\  
    \hline
    \end{tabular}
    }
  \end{minipage}
  ~
  \begin{minipage}{.53\linewidth}
    \centering
    \caption{Comparisons of PSNR with different learning paradigms.}
    \label{tab:paradigm}
    \resizebox{1\textwidth}{!}{
    \begin{tabular}{|l|l||cc|}
    \hline
    \multicolumn{1}{|l|}{Types} & Methods &  DAVIS ($\sigma{=}30$)  & Set8 ($\sigma{=}30$) \\ 
    \hline\hline
    Traditional     & VBM4D \cite{maggioni2012bm4d} & 31.65 & 30.00      \\ 
    Unsupervised learning   & UDVD \cite{sheth2021unsupervised} & 33.78 & 32.90      \\ 
    Self-supervised learning & MF2F \cite{dewil2021self} & 33.91 & 31.84    \\ 
    \hline
    Supervised learning  & \bf{Ours} &  \bf{37.10} & \bf{33.87}  \\  
    \hline
    \end{tabular}
    }
  \end{minipage}
  \vspace{-3mm}
\end{table*}

\vspace{-1mm}
\subsection{Ablation Studies}
\vspace{-1mm}
\noindent\textbf{Effectiveness of each module.}
We investigate the effectiveness of each module in Table \ref{tab:ablation}.
Specifically, we conduct experiments by removing these modules.
The model without these modules has a performance drop, which demonstrates the importance of each module.
These results demonstrate the importance of our proposed flow refinement, multiscale architecture, mutual alignment and FGDP.

\vspace{0.5mm}
\noindent\textbf{Ablation study on noise types.}
To investigate the effectiveness of the noise types, we remove one noise degradation and compare the performance on our synthesized REDS4 with fixed blur and noise degradations. Here, we mainly consider blur degradations, camera noises, and video compression noises which usually require temporal information for better denoising. From Table \ref{tab:ablation_type}, training without any kind of noise leads to inferior performance, which demonstrates the dominant role.

\noindent\textbf{Efficiency.}
In Figure \ref{fig:diff} (d), we also compare the model size and runtime across different methods to evaluate the efficiency of networks.
Our model achieves the best performance gains with a similar model size and runtime.
In particular, for the largest noise level of 50, our model outperforms VRT \cite{liang2022vrt} with smaller model size and faster inference time. 
Our model yields a PSNR improvement of \textbf{0.72dB}.

\begin{figure}[t]
  \begin{center}
  \includegraphics[width=1\linewidth]{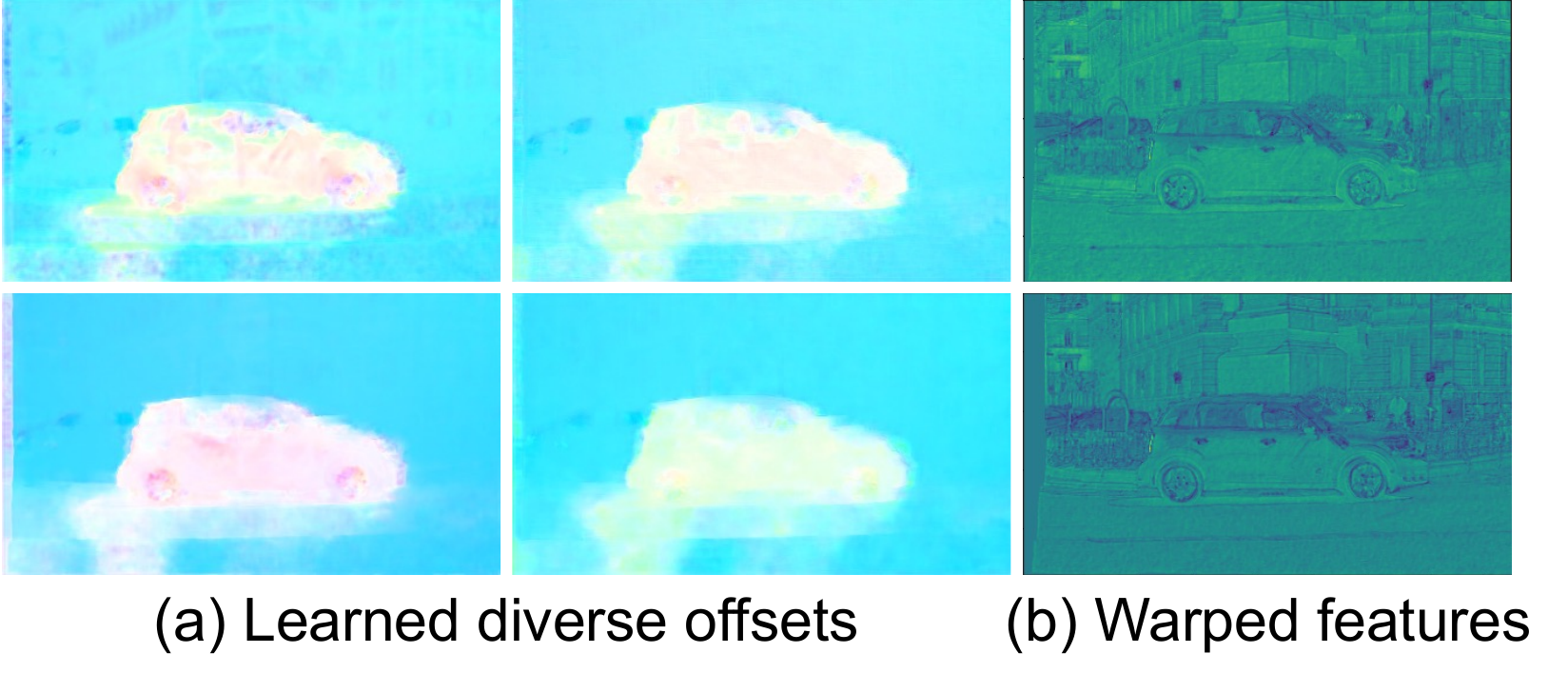}
  \end{center}
  \vspace{-5mm}
  \caption{Results of flows, learned offsets and warped features. }
  \label{fig:diverse_flow}
  \vspace{-5mm}
\end{figure}

\vspace{0.5mm}
\noindent\textbf{Results of clipped noise image denoising.}
We also train non-blind and blind models on clipped AWGN of DAVIS. 
In Table~\ref{tab:clip}, our model obtains the best performance under different noise levels.
In addition, Table \ref{tab:comp_img} shows that our method has better performance than image restoration methods since our
 model exploits the temporal information.

\vspace{0.5mm}
\noindent\textbf{Diversity of learned offsets.}
We further visualize the diversity of the offsets and aligned features in Figure \ref{fig:diverse_flow}.
With the refined flow and our proposed architecture, the diversity and quality of the offsets in the feature alignment can be increased. 
It means that the offset diversity helps the feature alignment to learn complementary offsets. In this way, it can alleviate the noise-occlusion issue and reduce warping errors.
As a result, these offsets help yield better-warped features with more details during the feature propagation.

\vspace{0.5mm}
\noindent\textbf{Different learning paradigms.}
We also compare our method with unsupervised learning and self-supervised learning paradigms (\ie UDVD \cite{sheth2021unsupervised} and MF2F \cite{dewil2021self}) on DAVIS and Set8 datasets, as shown in Table \ref{tab:paradigm}. 
With the help of the ground-truth videos in the training, we are able to train a better model to map a noisy video to clean one.

\vspace{0.5mm}
\noindent\textbf{Comparisons of flows.}
The optical flows between the frames $\bI_t$ and $\bI_{t+2}$ can be calculated directly or in a second-order way ( the same as \cite{chan2021basicvsrpp}).
We compare these two ways and show the visual quality of flows in the supplementary.
The latter way causes the error will be accumulated in the propagation.
In contrast, directly calculating the flow with our flow refinement has better performance.

\noindent\textbf{Flow estimation robustness against noise.}
The proposed method leads to more robust flow estimation in the presence of different noise levels. More analysis is in the supplementary.

\noindent\textbf{Video visual quality comparison.}
Quality of video denoising results can be better evaluated in the form of videos instead of images. In the supplementary, we provide more videos for comparison.

\section{Conclusion}
\vspace{-1mm}
In this paper, we propose a simple-but-effective video denoising method which achieves state-of-the-art performance on both synthetic Gaussian denoising and real video denoising.
Our proposed method mainly contains a denoising-oriented flow refinement (DFR) module and a flow-guided mutual denoising propagation (FMDP) module.
The DFR module improves the robustness of the optical flow under different noise levels.
The FMDP module makes use of the improved optical flow to mutually guide the feature propagation and alignment in multiple scales.
Moreover, we design a new noise degradation model for the real-world video denoising task which considers different kinds of noise with a random shuffle.
In addition, we propose a new real video denoising dataset with a large range of scenes and different noises.
Our model has shown good generalization performance on unseen real videos.
Extensive experiments demonstrate the effectiveness and superiority of our denoising method.

{\small
\bibliographystyle{ieee_fullname}
\bibliography{egbib}
}

\newpage
\appendix

\begin{center}
    \Large{\textbf{Supplementary Materials}}
\end{center}

\maketitle
\ificcvfinal\thispagestyle{empty}\fi

\setcounter{section}{0}
\renewcommand\thesection{\Alph{section}}
\renewcommand\thefigure{S\arabic{figure}}    
\setcounter{figure}{0}    
\renewcommand\thetable{S\arabic{table}}    
\setcounter{table}{0}
\renewcommand{\theequation}{S\arabic{equation}}
\setcounter{equation}{0}

\def\red{\textcolor{red}}

\vspace{-5mm}
\paragraph{Organization.}
In Section \ref{suppsec:degradation_detail}, we provide detailed settings of our video noise degradations. 
In Section \ref{suppsec:exp}, we provide more experimental  details and additional results. It also includes the qualitative evaluation of the flow robustness against different noise levels~(\ref{subsec:noise}).  In Section~\ref{sec:flow}, we provide the flow comparison when trained in the first- or second-order fashion. 
In Section \ref{suppsec:limit}, we discuss the limitations and societal impacts of our proposed method.

\vspace{-1mm}
\section{Experiment Details of Noise Degradation} \label{suppsec:degradation_detail}

\vspace{-1mm}
\noindent\textbf{Noise.}
In the experiment, we consider 6 kinds of noises in the degradations, including Gaussian noise, Poisson noise, Speckle noise, Processed camera sensor noise, JPEG compression noise and video compression noise.

\begin{itemize}[leftmargin=*]
    \vspace{-0.2cm}
    \item \emph{Gaussian noise.}
    When there is no prior information on noise, one can add Gaussian noise into a video sequence.
    Given a clean video $\bI$, we synthesize a noisy video by
    \begin{align}
        g_1(\bI) = \bI + \bZ, 
    \end{align}
    where we sample the noise levels $\sigma$ in $\bZ$ from $[2, 50]$.
    \vspace{-0.2cm}
    \item \emph{Poisson noise.}
    In electronics, Poisson noise is a type of shot noise that occurs in photon counting in optical devices.
    Such noise arises from the discrete nature of the electric charge, and it can be modeled by a Poisson process. 
    Given a clean video $\bI$, we synthesize a noisy video by 
    \begin{align}
        g_2(\bI) = \bI + \bZ,
    \end{align}
    where $\bZ=\bZ'-\bI$ and $\bZ' \sim \mP(10^{\alpha}\cdot \bI)/10^{\alpha}$. 
    We add Poisson noise in color images by sampling different noise levels.
    We first multiply the clean video by $10^{\alpha}$ in the function of Poisson distribution, where ${\alpha}$ is uniformly chosen from $[2, 4]$ and divide by $10^{\alpha}$.
    \vspace{-0.1cm}
    \item \emph{Speckle noise.}
    Speckle noise exists in synthetic aperture radar (SAR), medical ultrasound and optical coherence tomography images.
    We simulate such noise by multiplying the clean image $\bx$ and Gaussian noise $\bZ$, \ie $\bI*\bZ$.
    Then, we synthesize noisy video by 
    \begin{align}
        g_3(\bI)=\bI+\bI*\bZ,
    \end{align}
    We sample the level of this noise from $[0, 50]$.
    \vspace{-0.2cm}
    \item \emph{Processed camera sensor noise.}
    In modern digital cameras, the processed camera sensor noise originates from image signal processing (ISP). 
    Inspired by \cite{zhang2022scunet}, the reverse ISP pipeline first gets the raw image from an RGB image, then the forward pipeline constructs a noisy raw image by adding noise to the raw image,
    which is denoted by 
    \begin{align}
        g_4(\bI)=\text{forward}(\text{reverse}(\bI)).\vspace{-0.5cm}
    \end{align}
    \item \emph{JPEG compression noise.}
    It is widely used to reduce the storage for digital images with fast encoding and decoding \cite{zhang2021designing}.
    Such JPEG compression often causes $8 \times 8$ blocking artifacts.
    The degree of blocking artifacts depends on the quality of compression.
    We synthesize frames by 
    \begin{align}
        g_5(\bI)=\text{Dec}(\text{Enc}(\bI)).
    \end{align}
    The JPEG quality factor is uniformly chosen from $[30, 95]$.
    \vspace{-0.2cm}
    \item \emph{Video compression noise.}
    Videos sometimes have compression artifacts and are present on videos encoded in different formats.
    We use Pythonic $\text{av}$ in FFmpeg, \ie 
    \begin{align}
        g_6(\bI){=}\text{av}(\bI).
    \end{align}
    We randomly selected codecs from [`libx264', `h264', `mpeg4'] and bitrate from [1e4, 1e5] during training.
\end{itemize}

\vspace{0.2cm}
\noindent\textbf{Blur.}
In addition to noise, most real-world videos inherently suffer from blurriness.
Thus, we consider two blur degradations, including Gaussian blur and resizing blur.
\begin{itemize}[leftmargin=*]
    \item \emph{Gaussian blur.}
    We synthesize Gaussian blur with different kernels, including [`iso', `aniso', `generalized\_iso', `generalized\_aniso', `plateau\_iso', `plateau\_aniso', `sinc'].
    We randomly choose these kernels with the probabilities of [0.405, 0.225, 0.108, 0.027, 0.108, 0.027, 0.1].
    The settings of this blur are the same as \cite{chan2021realbasicvsr}.
    \vspace{-0.1cm}
    \item \emph{Resizing blur.}
    We randomly draw the resize scales from [0.5, 2], and choose the interpolation mode from [`bilinear', `area', `bicubic'] with the same probability of $1/3$.
\end{itemize}

\section{More Experiments} \label{suppsec:exp}
\vspace{-1mm}
\subsection{More Details of Experiment Setting}
We adopt the Adam optimizer \cite{kingma2014adam} and Cosine Annealing scheme \cite{loshchilov2016sgdr} to decay the learning rate from 
$1{\times}10^{-4}$ to $10^{-7}$.
The patch size is $256{\times}256$, and the batch size is 8.
The number of input frames is 15. 
All experiments are implemented in PyTorch 1.9.1.
We train the denoising model on 8 A100 GPUs.
We use the pre-trained SPyNet \cite{ranjan2017optical} to  estimate the flow and the SPyNet is further finetuned during training.
We train our video denoiser with 150K iterations.
For the synthetic Gaussian denoising experiments, the learning rate of the generator is $1{\times}10^{-4}$.
For real-world video denoising, the learning rates of the generator and discriminator are set to $5{\times}10^{-5}$ and $1{\times}10^{-4}$.
For the architecture of the generator, we use 5 residual blocks in the RDB block, use 7 residual blocks in the FMDP block, and set the feature channel as 64.
The architectures of the offset estimation module and mask estimation module are the same as \cite{chan2021basicvsrpp}. 
The architecture of the discriminator is the same as Real-ESRGAN \cite{wang2021realesrgan}.
When training classic video denoising, we use Charbonnier loss \cite{charbonnier1994two} because of its stability and good performance.
For video denoising on AWGN noise and the CRVD indoor dataset, we use the loss $\mL = \mL_{\text{pix}} + \lambda_{\text{flow}} \mL_{\text{flow}}$.
In the experiment, we set $\lambda_{\text{flow}}=0.1$.
For real video denoising experiments, we first use Charbonnier loss to train a model, then we finetune the network by using the perceptual loss $\mL_{\text{pix}}$ \cite{johnson2016perceptual} and adversarial loss $\mL_{\text{adv}}$ \cite{goodfellow2014gan}, \ie
$\mL = \mL_{\text{pix}} + \lambda_1 \mL_{\text{per}} + \lambda_2 \mL_{\text{adv}}$,
where $\lambda_1=1$ and $\lambda_2=5\times 10^{-1}$.
Code will be made publicly available.

\subsection{Training Loss and PSNR}
\begin{figure}[h]
    \begin{center}
    \vspace{-4mm}
    \includegraphics[width=1.04\linewidth]{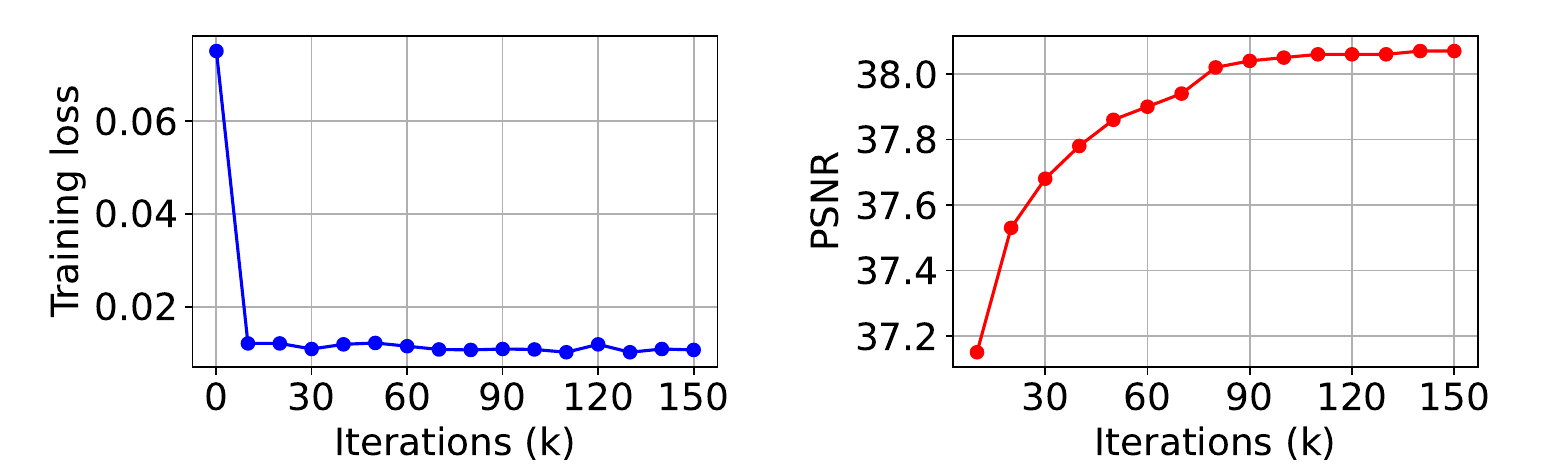}
    \end{center}
    \vspace{-5mm}
    \caption{An illustration of training loss and PSNR.}
    \label{fig:loss_psnr}
    \vspace{-0.5cm}
\end{figure}

To demonstrate the efficiency of our model, we show the training loss and PSNR validation, as shown in Figure \ref{fig:loss_psnr}.
At every 10K iterations, the PSNR value is calculated on Set8 with the noise level of 10. 
The total training iterations is 150k and takes 3 days.
The training loss decreases rapidly at early iterations and stays steady in the later iterations.
The PSNR values on Set8 increase during the training.
These results demonstrate that our model is easy to train to have good performance.

\subsection{Results on CRVD Outdoor Dataset}
\begin{figure}[h]
    \begin{center}
    \vspace{-3.5mm}
    \includegraphics[width=1.0\linewidth]{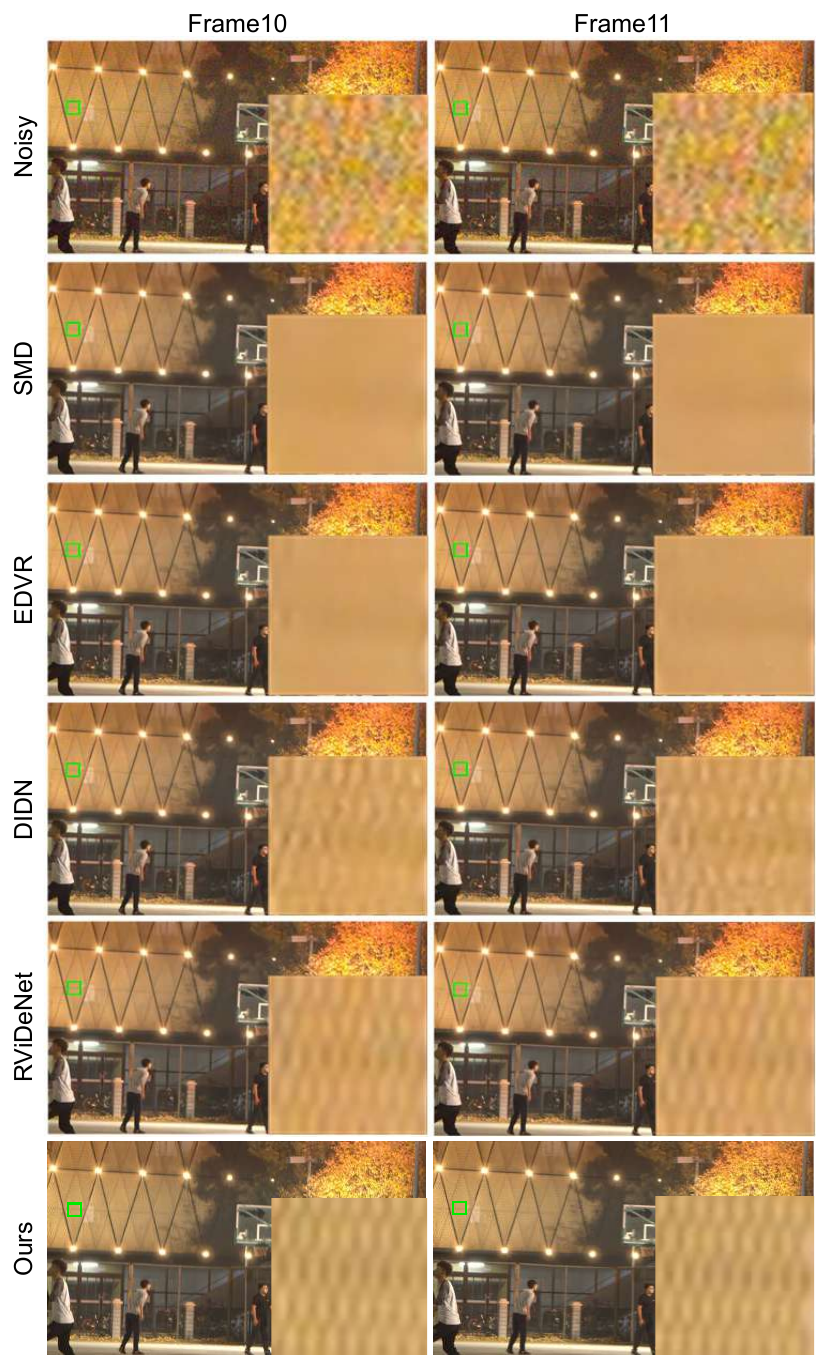}
    \end{center}
    \vspace{-3mm}
    \caption{Visual comparisons of different methods for two consecutive frames from the outdoor scene4 under ISO25600. These visual results come from \cite{yue2020rvidenet}.} 
    \label{fig:crvd_outdoor}
    \vspace{-0.3cm}
\end{figure}

In Figure \ref{fig:crvd_outdoor}, we compare different methods under ISO25600.
Our method has better visual quality than these methods.
Specifically, our method synthesizes a sharper grid than other methods. Please zoom in for a better observation of the figure.
In contrast, SMD \cite{chen2019smd}, DIDN \cite{yu2019didn} and EDVR \cite{wang2019edvr} have smooth textures and lose high-frequency details.

\subsection{Comparison on FLOPs}

Model inference time was provided in Figure \red{2} (d) in the paper, which can reflect the efficiency of the models. We compared the FLOPs and PSNR performance of different video denoising methods in Table \ref{supptab:comp_flops}. Here, the FLOPs is measured in TITAN RTX GPU with the spatial resolutions of $256{\times}256$. Our model achieves the best PSNR performance, although it has more FLOPs than BasicVSR++ \cite{chan2021basicvsrpp} due to the multi-scales. Besides, our model outperforms VRT \cite{liang2022vrt} with much fewer FLOPs. 

\begin{table}[h]
\vspace{-5pt}
\caption{Comparison with different methods on FLOPs.}
\label{supptab:comp_flops}
\centering
\resizebox{0.4\textwidth}{!}{
\begin{tabular}{|c||ccc|}
\hline
\multicolumn{1}{|c||}{Methods} & BasicVSR++ \cite{chan2021basicvsrpp} & VRT \cite{liang2022vrt} & Ours \\ 
\hline\hline
FLOPs (G) & 42.8  & 721.9 & 182.8 \\ 
~~~PSNR (dB)~~~ & 36.24 & 37.03 & 37.58 \\
\hline
\end{tabular}
}
\vspace{-5pt}
\end{table}

\begin{figure*}[t]
    \begin{center}
    \includegraphics[width=1.0\linewidth]{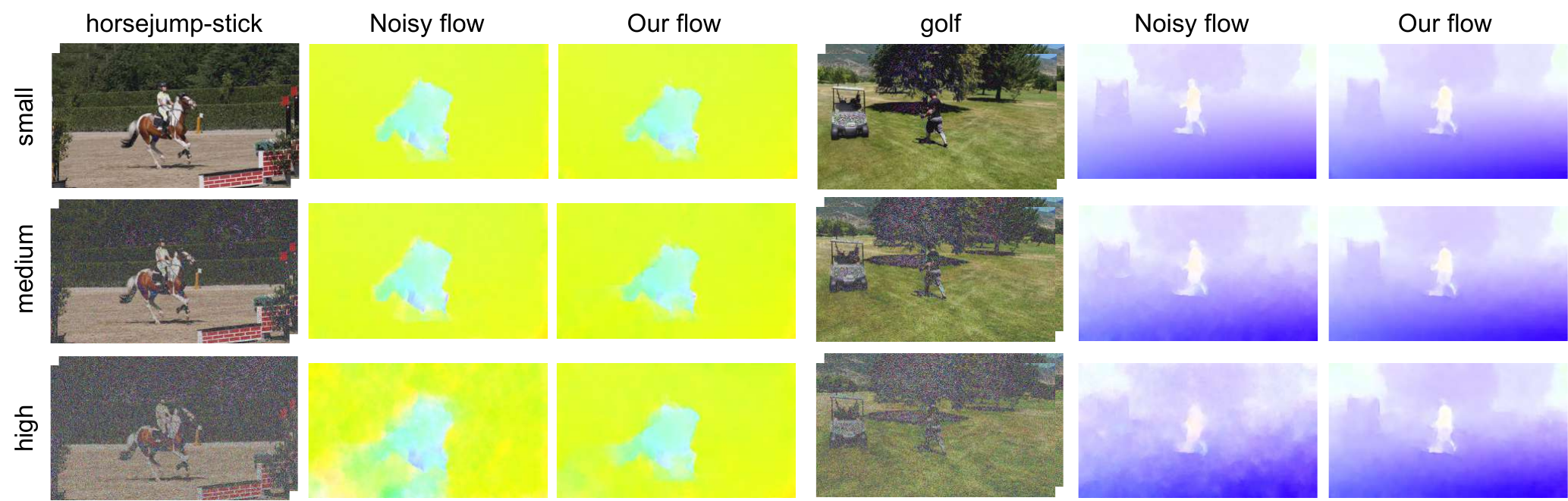}
    \end{center}
    \vspace{-3mm}
    \caption{Visualization of flows for different levels of noise on DAVIS.}
    \label{fig:robust_flow_noise}
\end{figure*}

\subsection{Generalization of Real Video Denoising Model}
To investigate the generalization performance, we test our method on REDS4 testing set with different noise types.
Specifically, we use REDS4 as a base and synthesize an additional noise type into the videos for each column in Table~\ref{supptab:generalization}. The trained model is then tested on the synthesized video with new noise. The different noise types are Gaussian noise, Poisson noise, Speckle noise, Camera noise, JPEG compression noise and Video compression noise. 
The levels of Gaussian and Speckle noise are 10, the scale of Poisson is 0.05, the quality scale of JPEG compression noise is 80, and the codec and bitrate of Video compression noise are `mpeg4' and $1e5$.
Our video denoiser has good generalization performance on other noise.

\begin{table}[h!]
  \caption{Generalization to different kinds of noise on REDS4.}
  \vspace{1mm}
  \label{supptab:generalization}
  \centering
  \resizebox{0.48\textwidth}{!}{
  \begin{tabular}{|c||cccccc|}
    \hline
    \multicolumn{1}{|c||}{Types} & \!\!\makecell{\scalebox{1}{Gaussian}\\noise}\!\! & \!\!\makecell{\scalebox{1}{Poisson}\\noise}\!\! & \!\!\makecell{\scalebox{1}{Speckle}\\noise}\!\! & \!\!\makecell{\scalebox{1}{Camera}\\noise}\!\! & \!\!\makecell{\scalebox{1}{JPEG}\\comp.}\!\! & \!\!\makecell{\scalebox{1}{Video}\\comp.}\!\!  \\
    \hline
    \hline
    \!\!Ours-real\!\! & {28.03} & 28.17 & {28.14} & {28.63} & {28.18} & {26.82} \\
    \hline
  \end{tabular}
  }
\end{table}

\subsection{More Details of Our RealNoise Videos}
To evaluate the generalization of real-world video denoising methods, it is important to collect a benchmark that covers a wide range of scenes and noises. 
Most existing datasets (\eg \cite{yue2020crvd}) are captured by one camera with different ISO settings.
However, such a dataset has a distribution mismatching gap compared with real-world noisy videos.
As a result, performing well in this kind of dataset may have poor generalization on videos in the wild.
To address this, we propose to capture the RealNoise dataset and select as many scenes and noises as possible. 
For example, the collected noisy videos include low-light, underwater, aerial videos, old films, weather, people, street, forest and natural scenery \etc.
Due to the file size limits, we only provide small video examples with the denoising results in the supplementary. The full dataset will be released. 

\begin{figure}[t]
  \begin{center}
  \vspace{-3mm}
  \includegraphics[width=1\linewidth]{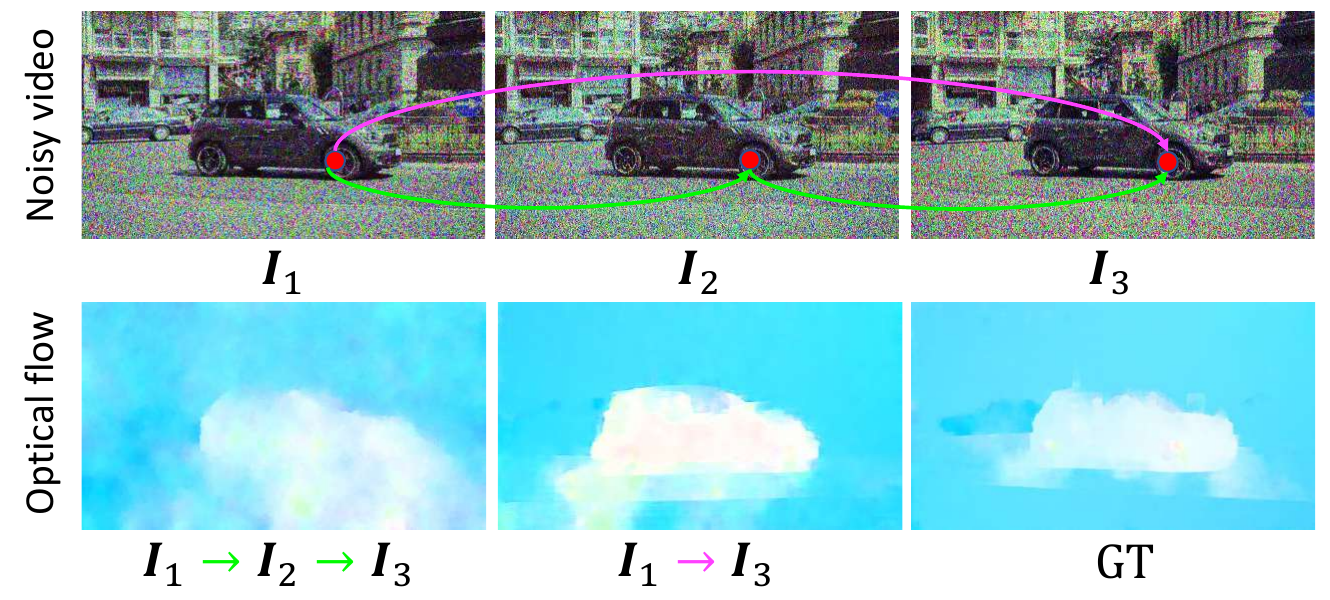}
  \end{center}
  \vspace{-4mm}
  \caption{Cumulative effect of the error of second-order flows.}
  \label{fig:flow1to2to3}
  \vspace{-3mm}
\end{figure}

\begin{table}[t]
  \caption{Comparison with different types of flows.}
  \vspace{1mm}
  \label{supptab:flow}
  \centering
  \resizebox{0.48\textwidth}{!}{
  \begin{tabular}{|c||ccc|}
    \hline
    \multicolumn{1}{|c||}{Types} & First-order & Second-order & Second-order with flow loss  \\
    \hline
    \hline
    PSNR & {34.45} & 34.90 & {35.08}  \\
    \hline
  \end{tabular}
  }
\end{table}

\begin{figure*}[t!]
	\centering
	\renewcommand{\g}{-0.7mm}
	\renewcommand{\tabcolsep}{1.8pt}
	\renewcommand{\arraystretch}{1}
	\resizebox{0.99\linewidth}{!} {
		\hspace{-4.5mm}
		\begin{tabular}{cc}
			\renewcommand{\name}{figures/set8/00000000_}
			\renewcommand{\h}{0.12}
			\renewcommand{\w}{0.2}
			\begin{tabular}{cc}
				\begin{adjustbox}{valign=t}
					\begin{tabular}{c}
		         	\includegraphics[trim={0 0 0 0},clip, width=0.354\textwidth]{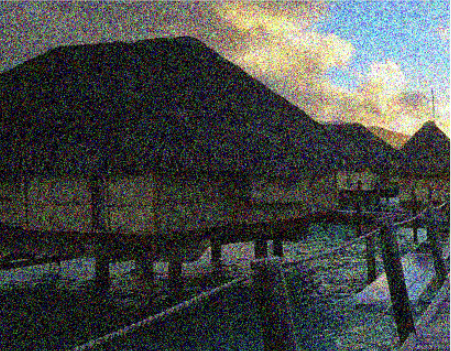}
						\\
						 00000000, hypersmooth, Set8
					\end{tabular}
				\end{adjustbox}
				\begin{adjustbox}{valign=t}
					\begin{tabular}{cccccc}
						\includegraphics[trim={0 0 0 0},clip,height=\h \textwidth, width=\w \textwidth]{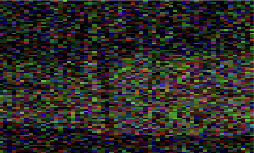} \hspace{\g} &
						\includegraphics[trim={0 0 0 0},clip,height=\h \textwidth, width=\w \textwidth]{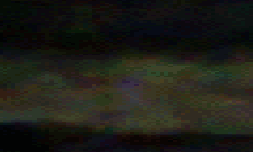} \hspace{\g} &
						\includegraphics[trim={0 0 0 0},clip,height=\h \textwidth, width=\w \textwidth]{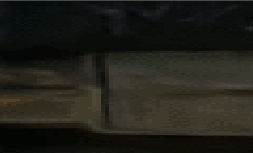} & 
						\includegraphics[trim={0 0 0 0},clip,height=\h \textwidth, width=\w \textwidth]{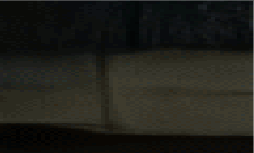} \hspace{\g} 
						\\
						\vspace{-0.5pt}Noisy frame ($\sigma{=}50$)\vspace{-4.5pt} &
						VBM4D \cite{maggioni2012bm4d} & DVDnet \cite{tassano2019dvdnet} & FastDVDnet \cite{tassano2020fastdvdnet}
						\\
						\vspace{-2mm}
						\\
						\includegraphics[trim={0 0 0 0},clip,height=\h \textwidth, width=\w \textwidth]{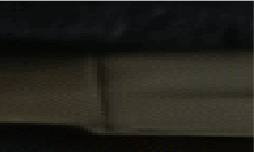} \hspace{\g} &
						\includegraphics[trim={0 0 0 0},clip,height=\h \textwidth, width=\w \textwidth]{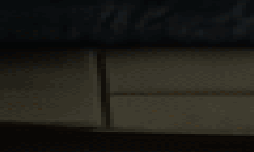} \hspace{\g} &
						\includegraphics[trim={0 0 0 0},clip,height=\h \textwidth, width=\w \textwidth]{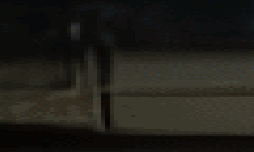}
						\hspace{\g} &		
						\includegraphics[trim={0 0 0 0},clip,height=\h \textwidth, width=\w \textwidth]{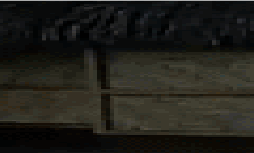}
						\\ 
						BasicVSR++ \cite{chan2021basicvsrpp} \hspace{\g} &	VRT \cite{liang2022vrt} \hspace{\g} &
						\textbf{Ours} &
						GT
						\\
					\end{tabular}
				\end{adjustbox}
			\end{tabular}
			
		\end{tabular}
	}
	\resizebox{0.99\linewidth}{!} {
		\hspace{-4.5mm}
		\begin{tabular}{cc}
			\renewcommand{\name}{figures/set8/00000060_}
			\renewcommand{\h}{0.12}
			\renewcommand{\w}{0.2}
			\begin{tabular}{cc}
				\begin{adjustbox}{valign=t}
					\begin{tabular}{c}
		         	\includegraphics[trim={0 0 0 0},clip, width=0.354\textwidth]{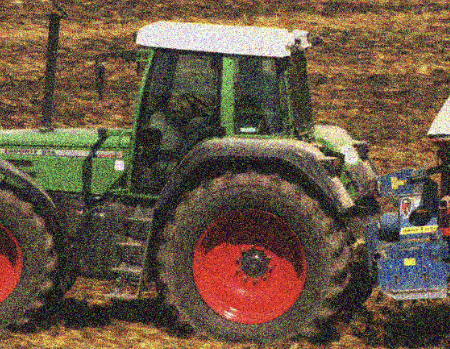}
						\\
						 00000060, tractor, Set8
					\end{tabular}
				\end{adjustbox}
				\begin{adjustbox}{valign=t}
					\begin{tabular}{cccccc}
						\includegraphics[trim={0 0 0 0},clip,height=\h \textwidth, width=\w \textwidth]{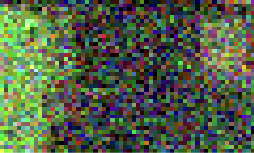} \hspace{\g} &
						\includegraphics[trim={0 0 0 0},clip,height=\h \textwidth, width=\w \textwidth]{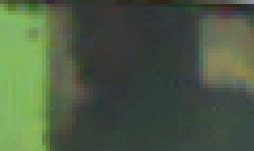} \hspace{\g} &
						\includegraphics[trim={0 0 0 0},clip,height=\h \textwidth, width=\w \textwidth]{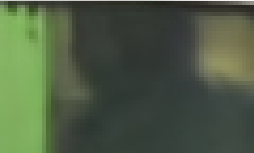} & 
						\includegraphics[trim={0 0 0 0},clip,height=\h \textwidth, width=\w \textwidth]{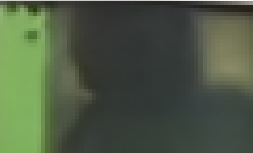} \hspace{\g} 
						\\
						\vspace{-0.5pt}Noisy frame ($\sigma{=}50$)\vspace{-4.5pt} &
						VBM4D \cite{maggioni2012bm4d} & DVDnet \cite{tassano2019dvdnet} & FastDVDnet \cite{tassano2020fastdvdnet}
						\\
						\vspace{-2mm}
						\\
						\includegraphics[trim={0 0 0 0},clip,height=\h \textwidth, width=\w \textwidth]{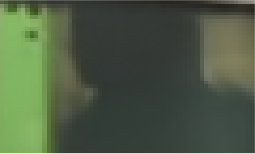} \hspace{\g} &
						\includegraphics[trim={0 0 0 0},clip,height=\h \textwidth, width=\w \textwidth]{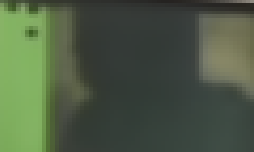} \hspace{\g} &
						\includegraphics[trim={0 0 0 0},clip,height=\h \textwidth, width=\w \textwidth]{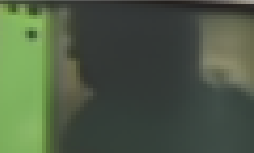}
						\hspace{\g} &		
						\includegraphics[trim={0 0 0 0},clip,height=\h \textwidth, width=\w \textwidth]{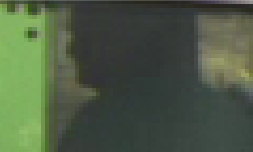}
						\\ 
						BasicVSR++ \cite{chan2021basicvsrpp} \hspace{\g} &	VRT \cite{liang2022vrt} \hspace{\g} &
						\textbf{Ours} &
						GT
						\\
					\end{tabular}
				\end{adjustbox}
			\end{tabular}
			
		\end{tabular}
	}
	\resizebox{0.99\linewidth}{!} {
		\hspace{-4.5mm}
		\begin{tabular}{cc}
			\renewcommand{\name}{figures/davis/00000011_}
			\renewcommand{\h}{0.12}
			\renewcommand{\w}{0.2}
			\begin{tabular}{cc}
				\begin{adjustbox}{valign=t}
					\begin{tabular}{c}
		         	\includegraphics[trim={0 0 0 0},clip, width=0.354\textwidth]{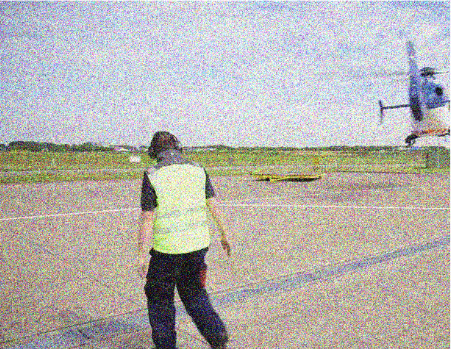}
						\\
						 00000011, helicopter, DAVIS 
					\end{tabular}
				\end{adjustbox}
				\begin{adjustbox}{valign=t}
					\begin{tabular}{cccccc}
						\includegraphics[trim={0 0 0 0},clip,height=\h \textwidth, width=\w \textwidth]{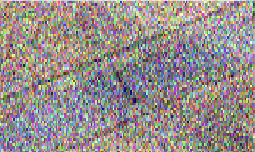} \hspace{\g} &
						\includegraphics[trim={0 0 0 0},clip,height=\h \textwidth, width=\w \textwidth]{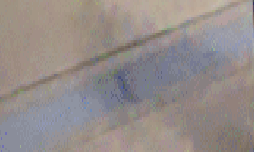} \hspace{\g} &
						\includegraphics[trim={0 0 0 0},clip,height=\h \textwidth, width=\w \textwidth]{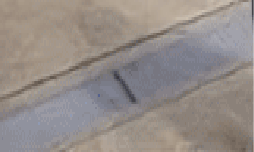} & 
						\includegraphics[trim={0 0 0 0},clip,height=\h \textwidth, width=\w \textwidth]{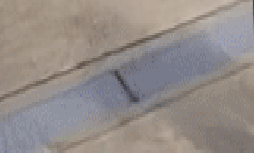} \hspace{\g} 
						\\
						\vspace{-0.5pt}Noisy frame ($\sigma{=}50$)\vspace{-4.5pt} &
						VBM4D \cite{maggioni2012bm4d} & DVDnet \cite{tassano2019dvdnet} & FastDVDnet \cite{tassano2020fastdvdnet}
						\\
						\vspace{-2mm}
						\\
						\includegraphics[trim={0 0 0 0},clip,height=\h \textwidth, width=\w \textwidth]{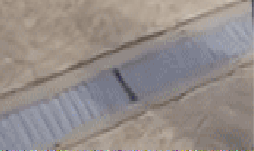} \hspace{\g} &
						\includegraphics[trim={0 0 0 0},clip,height=\h \textwidth, width=\w \textwidth]{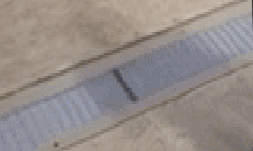} \hspace{\g} &
						\includegraphics[trim={0 0 0 0},clip,height=\h \textwidth, width=\w \textwidth]{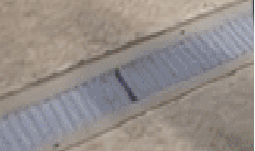}
						\hspace{\g} &		
						\includegraphics[trim={0 0 0 0},clip,height=\h \textwidth, width=\w \textwidth]{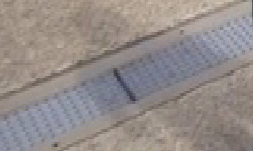}
						\\ 
						BasicVSR++ \cite{chan2021basicvsrpp} \hspace{\g} &	VRT \cite{liang2022vrt} \hspace{\g} &
						\textbf{Ours} &
						GT
						\\
					\end{tabular}
				\end{adjustbox}
			\end{tabular}
			
		\end{tabular}
	}
	\resizebox{0.99\linewidth}{!} {
		\hspace{-4.5mm}
		\begin{tabular}{cc}
			\renewcommand{\name}{figures/davis/00000073_}
			\renewcommand{\h}{0.12}
			\renewcommand{\w}{0.2}
			\begin{tabular}{cc}
				\begin{adjustbox}{valign=t}
					\begin{tabular}{c}
		         	\includegraphics[trim={0 0 0 0},clip, width=0.354\textwidth]{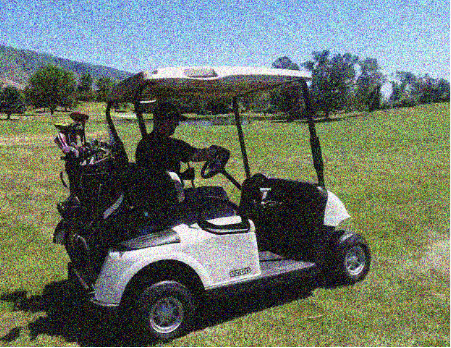}
						\\
						 00000073, golf, DAVIS 
					\end{tabular}
				\end{adjustbox}
				\begin{adjustbox}{valign=t}
					\begin{tabular}{cccccc}
						\includegraphics[trim={0 0 0 0},clip,height=\h \textwidth, width=\w \textwidth]{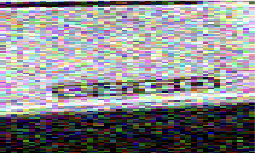} \hspace{\g} &
						\includegraphics[trim={0 0 0 0},clip,height=\h \textwidth, width=\w \textwidth]{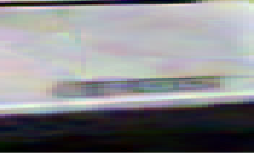} \hspace{\g} &
						\includegraphics[trim={0 0 0 0},clip,height=\h \textwidth, width=\w \textwidth]{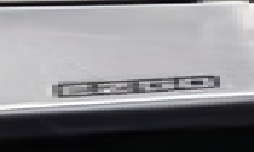} & 
						\includegraphics[trim={0 0 0 0},clip,height=\h \textwidth, width=\w \textwidth]{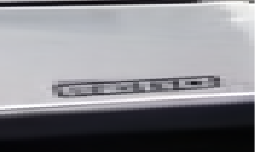} \hspace{\g} 
						\\
						\vspace{-0.5pt}Noisy frame ($\sigma{=}50$)\vspace{-4.5pt} &
						VBM4D \cite{maggioni2012bm4d} & DVDnet \cite{tassano2019dvdnet} & FastDVDnet \cite{tassano2020fastdvdnet}
						\\
						\vspace{-2mm}
						\\
						\includegraphics[trim={0 0 0 0},clip,height=\h \textwidth, width=\w \textwidth]{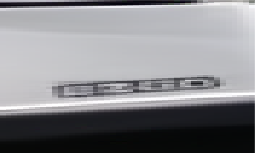} \hspace{\g} &
						\includegraphics[trim={0 0 0 0},clip,height=\h \textwidth, width=\w \textwidth]{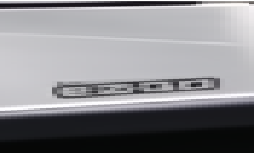} \hspace{\g} &
						\includegraphics[trim={0 0 0 0},clip,height=\h \textwidth, width=\w \textwidth]{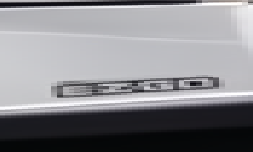}
						\hspace{\g} &		
						\includegraphics[trim={0 0 0 0},clip,height=\h \textwidth, width=\w \textwidth]{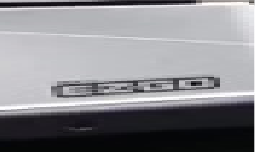}
						\\ 
						BasicVSR++ \cite{chan2021basicvsrpp} \hspace{\g} &	VRT \cite{liang2022vrt} \hspace{\g} &
						\textbf{Ours} &
						GT
						\\
					\end{tabular}
				\end{adjustbox}
			\end{tabular}
			
		\end{tabular}
	}
	\resizebox{0.99\linewidth}{!} {
		\hspace{-4.5mm}
		\begin{tabular}{cc}
			\renewcommand{\name}{figures/davis/00000000_}
			\renewcommand{\h}{0.12}
			\renewcommand{\w}{0.2}
			\begin{tabular}{cc}
				\begin{adjustbox}{valign=t}
					\begin{tabular}{c}
		         	\includegraphics[trim={0 0 0 0},clip, width=0.354\textwidth]{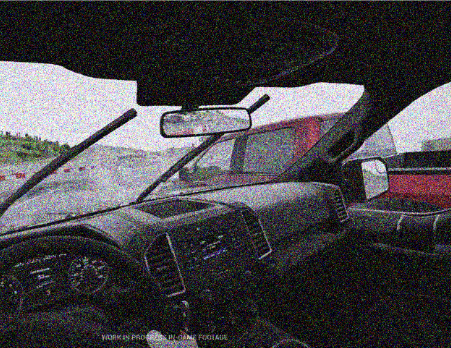}
						\\
						 00000000, car-race, DAVIS 
					\end{tabular}
				\end{adjustbox}
				\begin{adjustbox}{valign=t}
					\begin{tabular}{cccccc}
						\includegraphics[trim={0 0 0 0},clip,height=\h \textwidth, width=\w \textwidth]{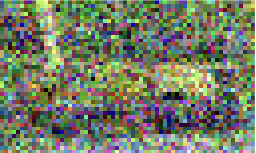} \hspace{\g} &
						\includegraphics[trim={0 0 0 0},clip,height=\h \textwidth, width=\w \textwidth]{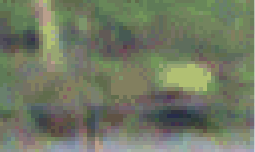} \hspace{\g} &
						\includegraphics[trim={0 0 0 0},clip,height=\h \textwidth, width=\w \textwidth]{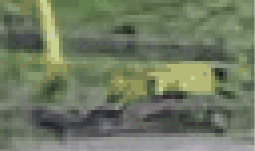} & 
						\includegraphics[trim={0 0 0 0},clip,height=\h \textwidth, width=\w \textwidth]{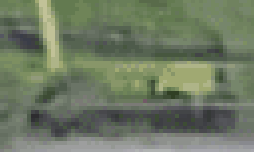} \hspace{\g} 
						\\
						\vspace{-0.5pt}Noisy frame ($\sigma{=}50$)\vspace{-4.5pt} &
						VBM4D \cite{maggioni2012bm4d} & DVDnet \cite{tassano2019dvdnet} & FastDVDnet \cite{tassano2020fastdvdnet}
						\\
						\vspace{-2mm}
						\\
						\includegraphics[trim={0 0 0 0},clip,height=\h \textwidth, width=\w \textwidth]{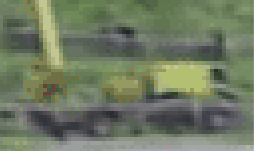} \hspace{\g} &
						\includegraphics[trim={0 0 0 0},clip,height=\h \textwidth, width=\w \textwidth]{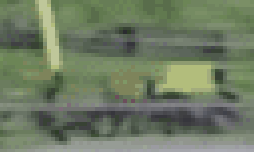} \hspace{\g} &
						\includegraphics[trim={0 0 0 0},clip,height=\h \textwidth, width=\w \textwidth]{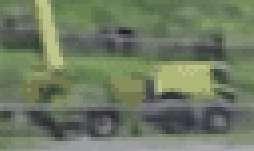}
						\hspace{\g} &		
						\includegraphics[trim={0 0 0 0},clip,height=\h \textwidth, width=\w \textwidth]{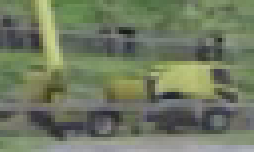}
						\\ 
						BasicVSR++ \cite{chan2021basicvsrpp} \hspace{\g} &	VRT \cite{liang2022vrt} \hspace{\g} &
						\textbf{Ours} &
						GT
						\\
					\end{tabular}
				\end{adjustbox}
			\end{tabular}
			
		\end{tabular}
	}
	\caption{Visual comparison of different methods on DAVIS under the noise level of 50.} 
	\label{suppfig:sr_quali}
\end{figure*}

\begin{figure*}[t]
	\centering
	\renewcommand{\g}{-0.7mm}
	\renewcommand{\tabcolsep}{1.8pt}
	\renewcommand{\arraystretch}{1}
	\resizebox{0.99\linewidth}{!} {
	\hspace{-4.5mm}
	\begin{tabular}{cc}
			\renewcommand{\name}{figures/ourdata/00000006_}
			\renewcommand{\h}{0.12}
			\renewcommand{\w}{0.2}
			\begin{tabular}{cc}
				\begin{adjustbox}{valign=t}
					\begin{tabular}{c}
		         	\includegraphics[trim={0 0 0 0},clip, width=0.354\textwidth]{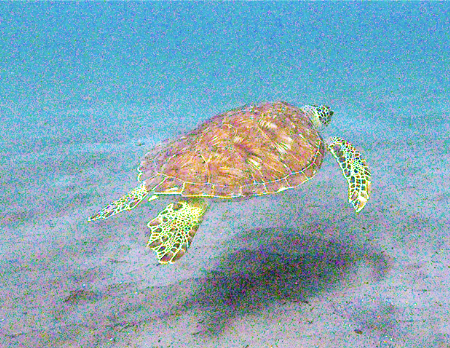}
						\\
						 RealNoise, underwater, frame6
					\end{tabular}
				\end{adjustbox}
				\begin{adjustbox}{valign=t}
					\begin{tabular}{cccccc}
						\includegraphics[trim={0 0 0 0},clip,height=\h \textwidth, width=\w \textwidth]{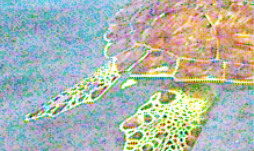} \hspace{\g} &
						\includegraphics[trim={0 0 0 0},clip,height=\h \textwidth, width=\w \textwidth]{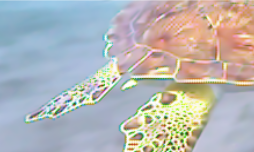} \hspace{\g} &
						\includegraphics[trim={0 0 0 0},clip,height=\h \textwidth, width=\w \textwidth]{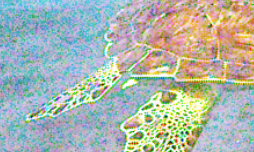} & 
						\includegraphics[trim={0 0 0 0},clip,height=\h \textwidth, width=\w \textwidth]{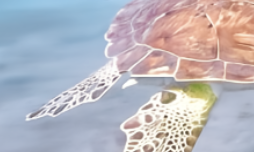} \hspace{\g} 
						\\
						Noisy frame \vspace{-0.5pt}\vspace{-4.5pt} &
						VBM4D \cite{maggioni2012bm4d} & Restormer \cite{zamir2021restormer} & SCUNet \cite{zhang2022scunet}
						\\
						\vspace{-2mm}
						\\
						\includegraphics[trim={0 0 0 0},clip,height=\h \textwidth, width=\w \textwidth]{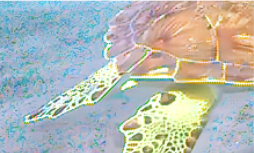} \hspace{\g} &
						\includegraphics[trim={0 0 0 0},clip,height=\h \textwidth, width=\w \textwidth]{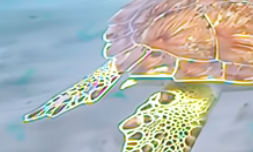} \hspace{\g} &
						\includegraphics[trim={0 0 0 0},clip,height=\h \textwidth, width=\w \textwidth]{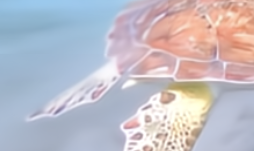}
						\hspace{\g} &		
						\includegraphics[trim={0 0 0 0},clip,height=\h \textwidth, width=\w \textwidth]{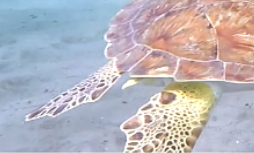}   
						\\ 
						ViDeNN \cite{claus2019videnn} \vspace{-0.5pt} \vspace{-4.5pt} & FloRNN \cite{li2022flornn}  &
						BasicVSR++ \cite{chan2022basicvsrpp2} & Ours \vspace{2mm}
						\\
					\end{tabular}
				\end{adjustbox}
			\end{tabular}
			
		\end{tabular}
	}
    \resizebox{0.99\linewidth}{!} {
	\hspace{-4.5mm}
	\begin{tabular}{cc}
			\renewcommand{\name}{figures/ourdata/00000050_}
			\renewcommand{\h}{0.12}
			\renewcommand{\w}{0.2}
			\begin{tabular}{cc}
				\begin{adjustbox}{valign=t}
					\begin{tabular}{c}
		         	\includegraphics[trim={0 0 0 0},clip, width=0.354\textwidth]{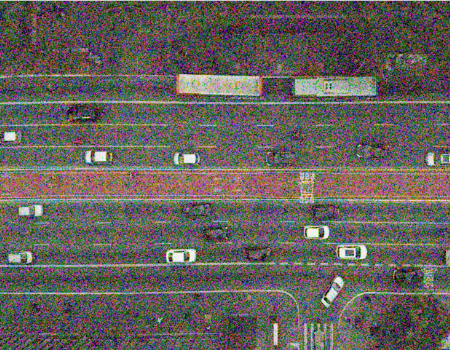}
						\\
						 RealNoise, car, frame50
					\end{tabular}
				\end{adjustbox}
				\begin{adjustbox}{valign=t}
					\begin{tabular}{cccccc}
						\includegraphics[trim={0 0 0 0},clip,height=\h \textwidth, width=\w \textwidth]{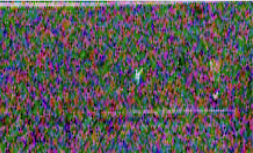} \hspace{\g} &
						\includegraphics[trim={0 0 0 0},clip,height=\h \textwidth, width=\w \textwidth]{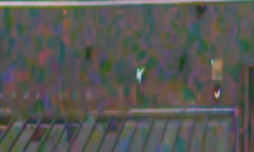} \hspace{\g} &
						\includegraphics[trim={0 0 0 0},clip,height=\h \textwidth, width=\w \textwidth]{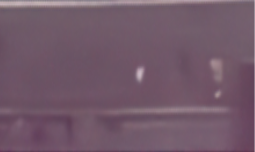} & 
						\includegraphics[trim={0 0 0 0},clip,height=\h \textwidth, width=\w \textwidth]{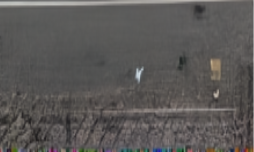} \hspace{\g} 
						\\
						Noisy frame \vspace{-0.5pt}\vspace{-4.5pt} &
						VBM4D \cite{maggioni2012bm4d} & Restormer \cite{zamir2021restormer} & SCUNet \cite{zhang2022scunet}
						\\
						\vspace{-2mm}
						\\
						\includegraphics[trim={0 0 0 0},clip,height=\h \textwidth, width=\w \textwidth]{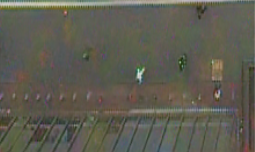} \hspace{\g} &
						\includegraphics[trim={0 0 0 0},clip,height=\h \textwidth, width=\w \textwidth]{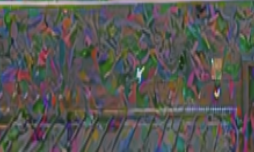} \hspace{\g} &
						\includegraphics[trim={0 0 0 0},clip,height=\h \textwidth, width=\w \textwidth]{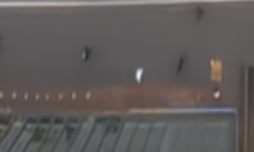}
						\hspace{\g} &		
						\includegraphics[trim={0 0 0 0},clip,height=\h \textwidth, width=\w \textwidth]{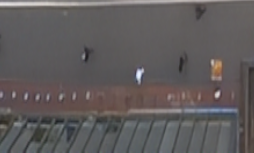}   
						\\ 
						ViDeNN \cite{claus2019videnn} \vspace{-0.5pt} \vspace{-4.5pt} & FloRNN \cite{li2022flornn}  &
						BasicVSR++ \cite{chan2022basicvsrpp2} & Ours \vspace{2mm}
						\\
					\end{tabular}
				\end{adjustbox}
			\end{tabular}
			
		\end{tabular}
	}
    \resizebox{0.99\linewidth}{!} {
	\hspace{-4.5mm}
	\begin{tabular}{cc}
			\renewcommand{\name}{figures/ourdata/00000001_}
			\renewcommand{\h}{0.12}
			\renewcommand{\w}{0.2}
			\begin{tabular}{cc}
				\begin{adjustbox}{valign=t}
					\begin{tabular}{c}
		         	\includegraphics[trim={0 0 0 0},clip, width=0.354\textwidth]{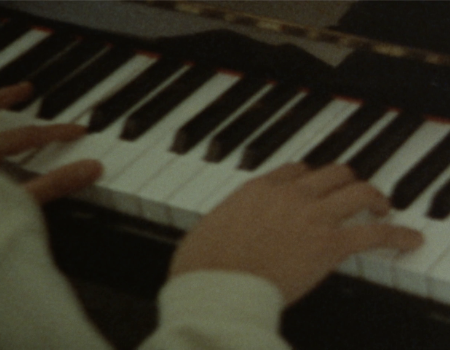}
						\\
						 RealNoise, old\_video, frame1
					\end{tabular}
				\end{adjustbox}
				\begin{adjustbox}{valign=t}
					\begin{tabular}{cccccc}
						\includegraphics[trim={0 0 0 0},clip,height=\h \textwidth, width=\w \textwidth]{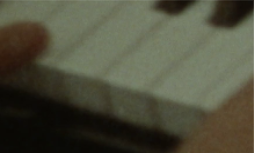} \hspace{\g} &
						\includegraphics[trim={0 0 0 0},clip,height=\h \textwidth, width=\w \textwidth]{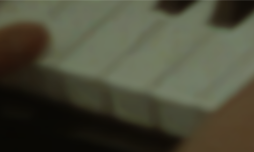} \hspace{\g} &
						\includegraphics[trim={0 0 0 0},clip,height=\h \textwidth, width=\w \textwidth]{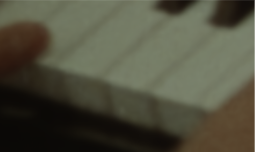} & 
						\includegraphics[trim={0 0 0 0},clip,height=\h \textwidth, width=\w \textwidth]{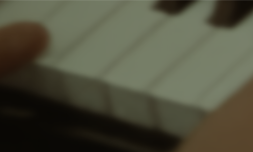} \hspace{\g} 
						\\
						Noisy frame \vspace{-0.5pt}\vspace{-4.5pt} &
						VBM4D \cite{maggioni2012bm4d} & Restormer \cite{zamir2021restormer} & SCUNet \cite{zhang2022scunet}
						\\
						\vspace{-2mm}
						\\
						\includegraphics[trim={0 0 0 0},clip,height=\h \textwidth, width=\w \textwidth]{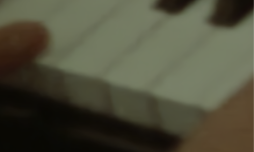} \hspace{\g} &
						\includegraphics[trim={0 0 0 0},clip,height=\h \textwidth, width=\w \textwidth]{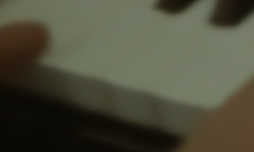} \hspace{\g} &
						\includegraphics[trim={0 0 0 0},clip,height=\h \textwidth, width=\w \textwidth]{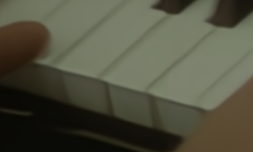}
						\hspace{\g} &		
						\includegraphics[trim={0 0 0 0},clip,height=\h \textwidth, width=\w \textwidth]{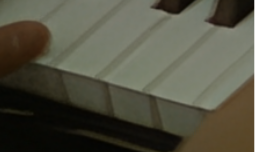}   
						\\ 
						ViDeNN \cite{claus2019videnn} \vspace{-0.5pt} \vspace{-4.5pt} & FloRNN \cite{li2022flornn}  &
						BasicVSR++ \cite{chan2022basicvsrpp2} & Ours \vspace{2mm}
						\\
					\end{tabular}
				\end{adjustbox}
			\end{tabular}
			
		\end{tabular}
	}
    \resizebox{0.99\linewidth}{!} {
	\hspace{-4.5mm}
	\begin{tabular}{cc}
			\renewcommand{\name}{figures/ourdata/00000005_}
			\renewcommand{\h}{0.12}
			\renewcommand{\w}{0.2}
			\begin{tabular}{cc}
				\begin{adjustbox}{valign=t}
					\begin{tabular}{c}
		         	\includegraphics[trim={0 0 0 0},clip, width=0.354\textwidth]{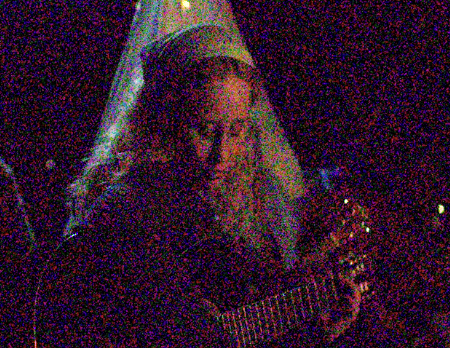}
						\\
						 RealNoise, people, frame5
					\end{tabular}
				\end{adjustbox}
				\begin{adjustbox}{valign=t}
					\begin{tabular}{cccccc}
						\includegraphics[trim={0 0 0 0},clip,height=\h \textwidth, width=\w \textwidth]{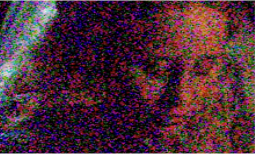} \hspace{\g} &
						\includegraphics[trim={0 0 0 0},clip,height=\h \textwidth, width=\w \textwidth]{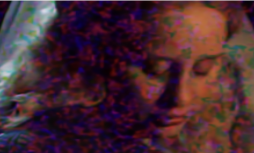} \hspace{\g} &
						\includegraphics[trim={0 0 0 0},clip,height=\h \textwidth, width=\w \textwidth]{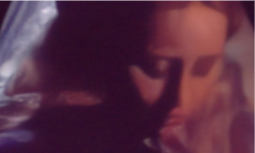} & 
						\includegraphics[trim={0 0 0 0},clip,height=\h \textwidth, width=\w \textwidth]{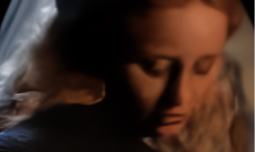} \hspace{\g} 
						\\
						Noisy frame \vspace{-0.5pt}\vspace{-4.5pt} &
						VBM4D \cite{maggioni2012bm4d} & Restormer \cite{zamir2021restormer} & SCUNet \cite{zhang2022scunet}
						\\
						\vspace{-2mm}
						\\
						\includegraphics[trim={0 0 0 0},clip,height=\h \textwidth, width=\w \textwidth]{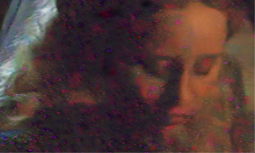} \hspace{\g} &
						\includegraphics[trim={0 0 0 0},clip,height=\h \textwidth, width=\w \textwidth]{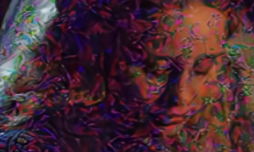} \hspace{\g} &
						\includegraphics[trim={0 0 0 0},clip,height=\h \textwidth, width=\w \textwidth]{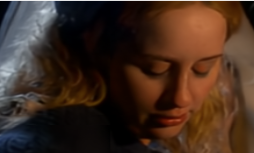}
						\hspace{\g} &		
						\includegraphics[trim={0 0 0 0},clip,height=\h \textwidth, width=\w \textwidth]{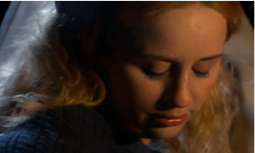}   
						\\ 
						ViDeNN \cite{claus2019videnn} \vspace{-0.5pt} \vspace{-4.5pt} & FloRNN \cite{li2022flornn}  &
						BasicVSR++ \cite{chan2022basicvsrpp2} & Ours \vspace{2mm}
						\\
					\end{tabular}
				\end{adjustbox}
			\end{tabular}
			
		\end{tabular}
	}
    \resizebox{0.99\linewidth}{!} {
	\hspace{-4.5mm}
	\begin{tabular}{cc}
			\renewcommand{\name}{figures/ourdata/00000009_}
			\renewcommand{\h}{0.12}
			\renewcommand{\w}{0.2}
			\begin{tabular}{cc}
				\begin{adjustbox}{valign=t}
					\begin{tabular}{c}
		         	\includegraphics[trim={0 0 0 0},clip, width=0.354\textwidth]{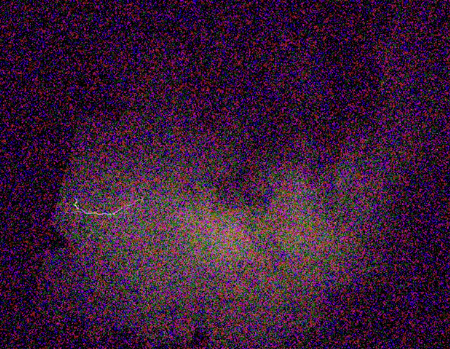}
						\\
						 RealNoise, thunder, frame9
					\end{tabular}
				\end{adjustbox}
				\begin{adjustbox}{valign=t}
					\begin{tabular}{cccccc}
						\includegraphics[trim={0 0 0 0},clip,height=\h \textwidth, width=\w \textwidth]{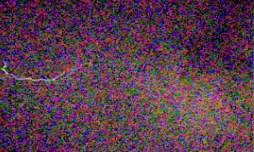} \hspace{\g} &
						\includegraphics[trim={0 0 0 0},clip,height=\h \textwidth, width=\w \textwidth]{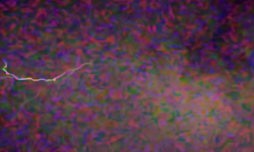} \hspace{\g} &
						\includegraphics[trim={0 0 0 0},clip,height=\h \textwidth, width=\w \textwidth]{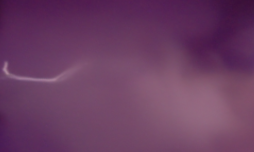} & 
						\includegraphics[trim={0 0 0 0},clip,height=\h \textwidth, width=\w \textwidth]{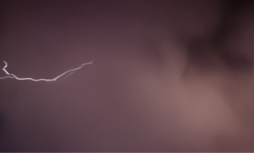} \hspace{\g} 
						\\
						Noisy frame \vspace{-0.5pt}\vspace{-4.5pt} &
						VBM4D \cite{maggioni2012bm4d} & Restormer \cite{zamir2021restormer} & SCUNet \cite{zhang2022scunet}
						\\
						\vspace{-2mm}
						\\
						\includegraphics[trim={0 0 0 0},clip,height=\h \textwidth, width=\w \textwidth]{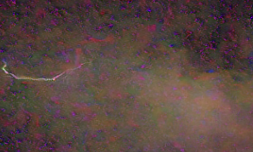} \hspace{\g} &
						\includegraphics[trim={0 0 0 0},clip,height=\h \textwidth, width=\w \textwidth]{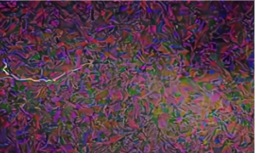} \hspace{\g} &
						\includegraphics[trim={0 0 0 0},clip,height=\h \textwidth, width=\w \textwidth]{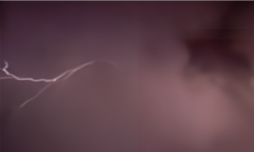}
						\hspace{\g} &		
						\includegraphics[trim={0 0 0 0},clip,height=\h \textwidth, width=\w \textwidth]{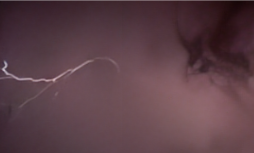}   
						\\ 
						ViDeNN \cite{claus2019videnn} \vspace{-0.5pt} \vspace{-4.5pt} & FloRNN \cite{li2022flornn}  &
						BasicVSR++ \cite{chan2022basicvsrpp2} & Ours \vspace{2mm}
						\\
					\end{tabular}
				\end{adjustbox}
			\end{tabular}
			
		\end{tabular}
	}
	\caption{Visual comparison of different methods on our proposed RealNoise dataset.} 
	\label{suppfig:realnoise_quali}
\end{figure*}

\subsection{More Qualitative Comparison}
In Figures \ref{suppfig:sr_quali} and \ref{suppfig:realnoise_quali}, we provide more visual comparisons for synthetic Gaussian denoising and real video denoising on our NoisyVideo dataset. 
Our model restores better structures and preserves a cleaner edge than previous state-of-the-art video denoising methods, even though the noise level is high. 
In particular, our model is able to synthesize the side profile in the second line of Figure \ref{suppfig:sr_quali}.
For real video denoising, our model achieves the best visual quality among different methods.
Specifically, our model can generate the stripe texture in the second example of Figure \ref{suppfig:realnoise_quali}.

\subsection{Flow Estimation Robustness Against Noise}\label{subsec:noise}
To investigate the robustness of our model to different kinds and levels of noise, we visualize the estimated flows in Figure \ref{fig:robust_flow_noise}.
We group the noise levels into small, medium and large by setting $\sigma{=}10, \sigma{=}30$ and $\sigma{=}50$, respectively.
For the small and medium noise levels, the noisy flow and our flow have similar results.
For the high noise level, our flow estimation model has a more robust performance.
The proposed method leads to more robust flow estimation in the presence of different noise levels.

\subsection{Comparisons with Flows} \label{sec:flow}

The optical flows between the frames $\bI_t$ and $\bI_{t+2}$ can be calculated directly or in a second-order way ( the same as \cite{chan2021basicvsrpp}).
We compare these two ways and show the visual quality of flows in Figure \ref{fig:flow1to2to3}.
The latter way causes the error will be accumulated in the propagation.
In contrast, directly calculating the flow with our flow refinement has better performance.

In addition, we compare three settings as follows.
(1) First-order flows mean that we estimate the optical flow of two adjacent frames, \ie $\bI_t{\to}\bI_{t+1}$ and $ \bI_t{\to}\bI_{t-1}$.
(2) Second-order flows mean that we estimate the optical flow of two adjacent frames and two frames apart, \ie $\bI_t{\to}\bI_{t+1}, \bI_t{\to}\bI_{t-1}, \bI_t{\to}\bI_{t+2}$ and $\bI_t{\to}\bI_{t-2}$.
(3) We also compare the second-order flows optimized with our proposed flow loss.
In Table \ref{supptab:flow}, we compare these types of flows on DAVIS under $\sigma=50$.
With the help of our flow refinement, using the second-order flows achieves the best performance.

\section{Limitations and Social Impacts} \label{suppsec:limit}
Our method achieves state-of-the-art performance in synthetic Gaussian denoising and real video denoising. 
This paper makes the first attempt to propose video noise degradations for real video denoising.
Our method can be used in some applications with positive social impacts.
For example, it can be used to restore old videos and remove compression noise from videos on the web.
However, there are some limitations in practice. 
First, it is hard for our model to remove blur artifacts which often occur in videos due to exposure time in different cameras.
However, our degradation pipeline mainly considers different kinds of noise. 
Second, it is challenging to remove big spot noise.
Third, our denoiser is trained with the GAN loss and it may change the identity of details (\eg human face) especially when the input is severely degraded.

\end{document}